\documentclass[sigconf, screen, nonacm]{acmart}

\usepackage{booktabs}
\usepackage{graphicx}
\usepackage{multirow}
\usepackage{amsmath}
\usepackage{xspace}
\usepackage{enumitem}
\usepackage{listings}
\usepackage{pgfplots}
\pgfplotsset{compat=1.18}
\usetikzlibrary{arrows.meta,calc,positioning,fit,backgrounds,
                 shadows.blur,decorations.pathreplacing}

\setcopyright{none}
\newcommand{\VDG}{\textsc{VDG}\xspace}

\title{Accuracy Without Grounding: Diagnosing Visual Dependency Dissociation in Video LLM Benchmarks}

\author{Jae Joong Lee}
\affiliation{
  \department{Department of Computer Science}
  \institution{Purdue University}
  \city{West Lafayette}
  \state{Indiana}
  \country{USA}
}

\begin{abstract}
Benchmark accuracy in video large language models (LLMs) is often treated as evidence of visual understanding. We audit this assumption across twenty models spanning 2--78B parameters and ten architecture families. We introduce the \textbf{Visual Dependency Gap} (\VDG), the difference in per-question correctness between original-video and black-screen conditions. Paired McNemar tests on MVBench show that accuracy and visual dependency are separable: models differ on original video ($p=0.0003$) but not on black screens ($p=0.53$). Across models, task-type rankings are stable: Attribute Perception is strongly visual, whereas Temporal Reasoning approaches the language-only baseline. A diagnostic ladder from black screen to single frame, shuffled frames, and original video reveals that frame diversity supplies most of the visual benefit, while temporal order contributes near-zero accuracy across sixteen open-weight models. An ablation from 0.5 to 24\,FPS rules out sparse sampling as the cause. H.264 experiments further show that stable aggregate accuracy conceals bidirectional question-level answer flips. The diagnostic also generalizes to four API-accessed models, whose \VDG values range from 0.025 to 0.315. These results motivate \VDG as a standard audit for whether video benchmarks measure visually grounded capability. Code is available at \url{https://github.com/JaeLee18/accuracy-without-grounding}.
\end{abstract}



\begin{document}

\maketitle

\section{Introduction}
\label{sec:intro}
\begin{figure*}[t]
    \centering
    \includegraphics[width=0.99\linewidth]{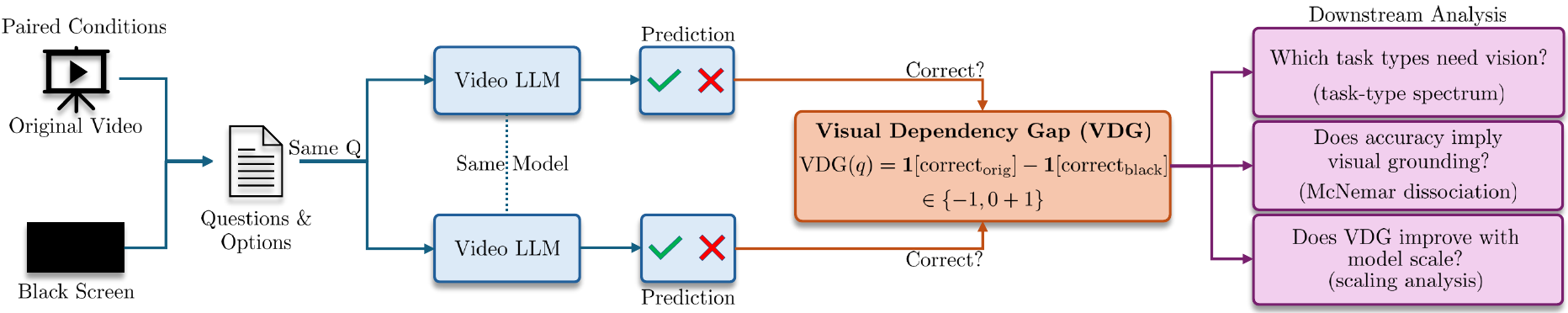}
    \caption{Each question is answered twice: once with the original video and once with a black screen. The per-question Visual Dependency Gap (VDG) (the difference in correctness between conditions) is aggregated to produce task-type spectra, McNemar dissociation tests, and scaling analyses across 20 models (2--78B, 10 families).}
    \label{fig:overview}
\end{figure*}

Video large language models (video LLMs)~\cite{maaz2023videochatgpt,li2023videochat,lin2023videollava,zhang2024llavavideo,cheng2024videollama2,tang2023videounderstanding} have emerged for multimodal understanding, extending the capabilities of foundation models~\cite{li2023blip2,dai2023instructblip,liu2024llava} to the spatio-temporal domain. From early systems like Video-ChatGPT~\cite{maaz2023videochatgpt} and VideoChat~\cite{li2023videochat} to recent architectures like Qwen2-VL~\cite{wang2024qwen2vl}, LLaVA-Video~\cite{zhang2024llavavideo}, InternVL2~\cite{chen2024internvl2}, VideoLLaMA2~\cite{cheng2024videollama2}, PLLaVA~\cite{chen2024pllava}, and proprietary systems~\cite{openai2023gpt4,reid2024gemini15}, these models are evaluated on standardized benchmarks~\cite{fu2024videomme,li2024mvbench,mangalam2023egoschema} whose leaderboard rankings shape architectural choices and deployment decisions across the field.

A core assumption of this evaluation method is that higher benchmark accuracy implies stronger visual understanding. If true, leaderboard improvements would reliably signal progress in visual reasoning. If false, the field risks optimizing for language-prior exploitation rather than genuine visual grounding\footnote{We use ``visual grounding'' to mean dependence of model predictions on visual input content, not the localization-specific usage common in referring expression literature. To avoid ambiguity, we primarily use the term ``visual dependency'' throughout.}, a distinction invisible to accuracy metrics alone. Testing this assumption is therefore essential to ensuring that benchmark progress translates to real-world visual capability.

Prior work has shown that visual QA benchmarks contain language-solvable questions: VQA-CP~\cite{agrawal2018vqacp} revealed answer-type priors, Chen et al.~\cite{chen2024mmstar} found image-free solvability in multimodal benchmarks, and TVBench~\cite{cores2024tvbench} showed that video LLM benchmarks are susceptible to text-only solving. However, these studies \emph{detect} the problem without quantifying it per-question, and none establishes that accuracy and visual dependency are statistically separable. The distinction matters: knowing that ``some questions are solvable without vision'' is weaker than proving that ``a more accurate model is not necessarily more visually dependent.''

We address this gap by introducing and validating the \textbf{Visual Dependency Gap} (\VDG), a per-question diagnostic that measures the difference in correctness between video-present and black-screen conditions (Figure~\ref{fig:overview}):
\[
\VDG(q) = \mathbf{1}[\text{correct at original}] - \mathbf{1}[\text{correct at black}] \;\in\; \{-1,\, 0,\, +1\}.
\]
Across twenty models spanning 2--78B parameters and ten architecture families (including proprietary API models), we show that benchmark accuracy and visual grounding are \emph{dissociable axes}: on MVBench, InternVL2-8B reliably outperforms Qwen2-VL-7B on original video ($p=0.0003$) yet the same pair is statistically indistinguishable on a black screen ($p=0.53$). The task-type \VDG spectrum is stable across all models, with Attribute Perception as the most visually grounded category ($\approx 0.41$) and Temporal Reasoning nearly indistinguishable from language-only performance ($\approx 0.06$). It is confirmed by an FPS ablation from 0.5 to 24\,FPS (Section~\ref{sec:fps}). A four-condition diagnostic ladder further reveals that temporal ordering contributes near-zero accuracy, while visual grounding is dominated by frame diversity (Section~\ref{sec:ladder}).

\noindent Our contributions are following:
\begin{enumerate}
  \item \textbf{McNemar dissociation} (Section~\ref{sec:dissociation}): paired McNemar testing proves that accuracy and visual dependency are \emph{statistically separable axes}, which is a stronger claim than prior work's ``models can answer without vision.''
  \item \textbf{Task-type taxonomy and cross-benchmark consistency} (Sections~\ref{sec:spectrum}--\ref{sec:validity}): a \VDG spectrum across six task types, consistent across twenty models and three benchmarks, where Video-MME labels predict MVBench classification at 100\% (7/7) and generalize to EgoSchema (zero source overlap).
  \item \textbf{Four-condition diagnostic ladder} (Section~\ref{sec:ladder}): decomposition of \VDG into spatial, frame-diversity, and temporal components across sixteen open-weight models reveals that temporal ordering contributes near-zero accuracy universally, and Qwen3-VL shows a diversity-extraction collapse relative to Qwen2.5-VL.
  \item \textbf{Compression robustness illusion} (Section~\ref{sec:crf}): flat CRF curves arise from bidirectional cancellation, and CRF-sensitive questions are disproportionately visually grounded, explaining why compression ``robustness'' is an artifact of benchmark composition.
\end{enumerate}

\section{Related Work}
\label{sec:related}

\noindent\textbf{Video LLM benchmarks.}
Video-MME~\cite{fu2024videomme}, MVBench~\cite{li2024mvbench}, and EgoSchema~\cite{mangalam2023egoschema} are the primary suites audited here, and the broader landscape includes ActivityNet-QA~\cite{yu2019activitynetqa}, NExT-QA~\cite{xiao2021nextqa}, SEED-Bench~\cite{li2024seedbench}, TempCompass~\cite{liu2024tempcompass}, LongVideoBench~\cite{wu2024longvideobench}, FunQA~\cite{xie2024funqa}, MMBench-Video~\cite{fang2024mmbenchvideo}, and MSRVTT-QA~\cite{xu2017msrvttqa}. Complementary image-domain benchmarks such as MMBench~\cite{liu2023mmbench}, MMMU~\cite{yue2024mmmu}, and MME~\cite{fu2024mme} have established evaluation standards that video benchmarks increasingly follow. Despite this proliferation, few benchmarks include diagnostics for whether questions require visual input. Buch et al.~\cite{buch2022revisiting} showed many video QA datasets are solvable from a single frame. Lei et al.~\cite{lei2023revealing} demonstrated single-frame bias in video-language learning, and Chen et al.~\cite{chen2024mmstar} found image-free solvability in multimodal benchmarks.

\noindent\textbf{Language priors and diagnostic baselines.}
Language priors in visual QA are well documented~\cite{goyal2017making,antol2015vqa,agrawal2018vqacp}, with subsequent work identifying unimodal shortcuts~\cite{cadene2019rubi,geirhos2020shortcut}, causal biases~\cite{niu2021counterfactual}, annotation artifacts~\cite{gururangan2018annotation}, and multimodal shortcut interactions~\cite{dancette2021beyond}. In the multimodal LLM era, Tong et al.~\cite{tong2024eyeswideshut} exposed systematic visual shortcomings, and Li et al.~\cite{li2023pope} proposed polling-based hallucination evaluation. For video, TVBench~\cite{cores2024tvbench} showed text-only solvability, and Lydakis et al.~\cite{lydakis2025videoimportant} analyzed training data composition, and MoReVQA~\cite{min2024morevqa} introduced LLM-only baselines. We show accuracy and grounding are \emph{dissociable axes} via paired McNemar testing, a stronger claim than prior detection-based approaches. Compression robustness has been studied for images~\cite{dodge2016understanding,hendrycks2019benchmarking} and under distribution shift~\cite{taori2020measuring}. We extend this to H.264 compression of video LLM benchmarks. Bowman and Dahl~\cite{bowman2021benchmarking} argued benchmarks need validity diagnostics beyond accuracy, and behavioral testing frameworks~\cite{ribeiro2020beyond} provide complementary methodology. Our \VDG addresses the specific validity concern of whether accuracy reflects visual grounding. Vision encoders range from CLIP~\cite{radford2021clip} and SigLIP~\cite{zhai2023sigmoid} to architecture-specific designs~\cite{dosovitskiy2021vit}. The \VDG spectrum is stable across all encoder types evaluated.

\noindent\textbf{Construct validity and diagnostic decomposition.}
Whether benchmark accuracy measures what it claims, a question of construct validity~\cite{schlangen2021targeting}, is increasingly examined in NLP and multimodal evaluation. Schlangen~\cite{schlangen2021targeting} argued that leaderboard scores lack interpretability without explicit construct definitions, and Raji et al.~\cite{raji2021ai} demonstrated that aggregate accuracy can conflate distinct capabilities into a single number that may not reflect any individual one. In the vision-language setting, Frank et al.~\cite{frank2021vision} probed cross-modal influence in multimodal transformers and found that model predictions often rely primarily on textual features even for nominally visual tasks. Y\"uksekg\"on\"ul et al.~\cite{yuksekgonul2023when} showed that many vision-language models behave as bags of words, failing to leverage compositional and spatial structure in either modality. Lee~\cite{lee2026lgip} introduced language-guided invariance probing to quantify whether VLMs are invariant to meaning-preserving paraphrases and sensitive to semantic flips, finding that even strong models such as SigLIP score corrupted captions above human descriptions. For video, the single-frame solvability documented by Buch et al.~\cite{buch2022revisiting} and Lei et al.~\cite{lei2023revealing} implies that temporal benchmarks may not require temporal reasoning, but these findings remain at the dataset level without per-question quantification. Our \VDG addresses this gap: it provides a per-question diagnostic of the accuracy--grounding dissociation, decomposes visual dependency into spatial, frame-diversity, and temporal components via a four-condition ladder, and establishes that the resulting task-type taxonomy generalizes across benchmarks, model families, and deployment conditions.

\section{Method}
\label{sec:method}

\subsection{Models and Benchmarks}

We evaluate twenty video LLMs spanning ten architecture families and 2--78B parameters, including both open-weight and proprietary API models. The three \emph{primary} models used for all experiments (VDG spectrum, McNemar, CRF, cross-benchmark):
\begin{itemize}
  \item \textbf{Qwen2-VL-7B-Instruct}~\cite{wang2024qwen2vl}: ViT~\cite{dosovitskiy2021vit} vision encoder with Qwen2 language model.
  \item \textbf{LLaVA-Video-7B}~\cite{zhang2024llavavideo}: CLIP~\cite{radford2021clip} ViT-L/14 with linear projection and Qwen2-7B language model.
  \item \textbf{InternVL2-8B}~\cite{chen2024internvl2}: InternViT-300M encoder with MLP connector and InternLM2, with no CLIP encoder.
\end{itemize}
For scale, generation, and ablation analyses (Sections~\ref{sec:fps}--\ref{sec:ladder}), we evaluate thirteen additional open-weight models in two precision tiers. \emph{Full-precision} (bf16): Qwen2-VL-2B, Qwen2.5-VL-3B/7B~\cite{bai2025qwen25vl}, Qwen3-VL-2B/8B~\cite{yang2025qwen3vl}, InternVL2-2B/26B, VideoLLaMA2-7B~\cite{cheng2024videollama2}. \emph{Extended} (4-bit NF4 quantization): Qwen2-VL-72B, Qwen2.5-VL-32B/72B, Qwen3-VL-32B, InternVL2-76B, InternVL2.5-78B. Four additional models are evaluated via API for \VDG generalization: GPT-4o-mini, Gemini~2.5 Flash Lite, Llama~3.2 11B Vision, and Nemotron Nano 12B~VL. Defended scaling claims are restricted to full-precision models, and 4-bit results are reported with quantization caveats (Section~\ref{sec:scale}).

We evaluate on three benchmarks:
\begin{itemize}
  \item \textbf{Video-MME}~\cite{fu2024videomme}: 600 questions stratified across 6 task types (100 each): OCR Problems, Action Recognition, Action Reasoning, Temporal Reasoning, Attribute Perception, Object Recognition. This 600-question stratified subset samples uniformly across all six task types to ensure equal statistical power per category. It is drawn from Video-MME's full question pool and used consistently across all four experimental conditions and all twenty models.
  \item \textbf{MVBench}~\cite{li2024mvbench}: 462 questions across 9 task types, including \texttt{action\_prediction}, \texttt{object\_existence}, \texttt{scene\_transition}, \texttt{state\_change}, \texttt{unexpected\_action}, \texttt{episodic\_reasoning}, and \texttt{action\_recognition}.
  \item \textbf{EgoSchema}~\cite{mangalam2023egoschema}: 500 five-choice questions over 3-minute egocentric videos from Ego4D~\cite{grauman2022ego4d}. EgoSchema has \emph{zero video source overlap} with Video-MME and MVBench, providing a fully independent test of whether \VDG-based tier predictions generalize beyond the Video-MME/MVBench construction family.
\end{itemize}

\subsection{Visual Dependency Gap}

For each question $q$, we define:
\begin{equation}
  \VDG(q) = \mathbf{1}[\text{correct}(q, \text{original})] - \mathbf{1}[\text{correct}(q, \text{black screen})]
  \label{eq:vdg}
\end{equation}
where the black-screen condition replaces all video frames with a solid black image while keeping the question text and answer options unchanged. $\VDG(q) \in \{-1, 0, +1\}$. Positive values indicate visual grounding (correct only with video), zero indicates language-prior sufficiency or universal failure, and negative values indicate that the visual stream is actively harmful.

The aggregate \VDG for a model over a set of questions $\mathcal{Q}$ is:
\[
  \overline{\VDG} = \frac{1}{|\mathcal{Q}|} \sum_{q \in \mathcal{Q}} \VDG(q) = \text{Acc}_\text{orig} - \text{Acc}_\text{black}
\]

While the aggregate $\overline{\VDG}$ reduces to an accuracy difference, the per-question formulation enables analyses the aggregate cannot: four-category decomposition identifying the 31\% ``pure visual core'' (Section~\ref{sec:leakage}), the diagnostic ladder decomposing \VDG into spatial, diversity, and temporal components (Section~\ref{sec:ladder}), and CRF-stratified sensitivity analysis (Section~\ref{sec:crf}). We validate \VDG through three channels: architecture-independent black-screen floors (Section~\ref{sec:convergence}), CRF-sensitivity enrichment, and cross-benchmark tier consistency (Section~\ref{sec:validity}).

\subsection{McNemar Dissociation Test}

To test whether two models differ in their dependence on the visual stream independently of their overall accuracy, we apply the McNemar test~\cite{mcnemar1947} to paired binary outcomes. Specifically, for model pair $(M_1, M_2)$ on condition $c \in \{\text{original}, \text{black screen}\}$, we test $H_0$ that the off-diagonal cells of the $2\times 2$ correctness contingency table are equal. A reliable result on the original condition indicates accuracy differences, while reliability on black screen indicates differences in language-prior exploitation. \emph{Dissociation} occurs when $p_\text{orig}$ and $p_\text{black}$ have opposite reliability patterns. All reported $p$-values survive Holm--Bonferroni correction for the six tests conducted across the two benchmarks (adjusted threshold for the smallest $p$: $0.05/6 = 0.0083$, with the smallest observed $p = 0.0003$).

\subsection{Compression Protocol}

Videos are re-encoded with H.264 at CRF $\in \{18, 23, 28, 33, 38\}$ using \texttt{ffmpeg}/\texttt{libx264}. CRF 18 is near-lossless (SSIM\,$=0.993$), while CRF 38 is heavily compressed (SSIM\,$=0.942$). As an external perceptual proxy, CLIP ViT-B/32 gives an embedding distance of 0.027 between original and CRF\,38 video (vs.\ 0.165 for Gaussian blur $\sigma=10$). Because none of the three primary models uses this exact encoder, we treat this value as a manipulation check, not as evidence of their internal encoder sensitivity.

\section{Results}
\label{sec:results}

\subsection{VDG Spectrum by Task Type}
\label{sec:spectrum}

\begin{figure}[t]
\centering
\begin{tikzpicture}
\begin{axis}[
    ybar, bar width=4pt, width=0.95\columnwidth, height=5cm,
    symbolic x coords={Attr.Perc,Obj.Rec,OCR,Act.Rec,Act.Reas,Temp.Reas},
    xtick=data, x tick label style={rotate=30,anchor=east,font=\footnotesize},
    ymin=0, ymax=0.6, ylabel={VDG}, ylabel style={font=\footnotesize},
    legend style={at={(0.99,0.99)},anchor=north east,font=\footnotesize,
                  legend image code/.code={\fill[##1] (0cm,-0.1cm) rectangle (0.25cm,0.1cm);}},
    legend columns=3,
    ymajorgrids=true, grid style=dashed,
    title style={font=\small},
    enlarge x limits=0.1,
]
\addplot[ybar, fill=blue!60, draw=blue!80] coordinates {
    (Attr.Perc,0.400)(Obj.Rec,0.250)(OCR,0.321)(Act.Rec,0.210)(Act.Reas,0.225)(Temp.Reas,0.000)
};
\addplot[ybar, fill=red!60, draw=red!80] coordinates {
    (Attr.Perc,0.460)(Obj.Rec,0.480)(OCR,0.330)(Act.Rec,0.320)(Act.Reas,0.160)(Temp.Reas,0.080)
};
\addplot[ybar, fill=green!60, draw=green!80] coordinates {
    (Attr.Perc,0.380)(Obj.Rec,0.380)(OCR,0.230)(Act.Rec,0.250)(Act.Reas,0.180)(Temp.Reas,0.090)
};
\legend{Qwen2-VL,LLaVA-Video,InternVL2}
\end{axis}
\end{tikzpicture}
\caption{Visual Dependency Gap (VDG) by task type on Video-MME. Temporal Reasoning shows VDG $\approx$ 0 for all three architectures, while Attribute Perception shows consistently high VDG. The ranking is stable across three architecturally distinct models (mean pairwise $r = 0.789$, individual pairs range from $p>0.05$ to $p=0.008$ at $k=6$, see text).}
\label{fig:vgg_spectrum}
\end{figure}
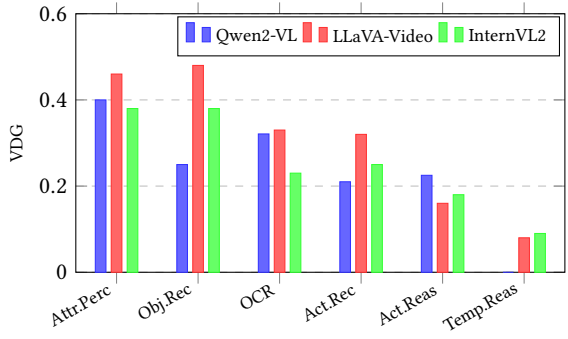

Table~\ref{tab:vgg_spectrum} reports per-task-type aggregate \VDG for all three models on Video-MME, along with bootstrap 95\% confidence intervals. Figure~\ref{fig:vgg_spectrum} visualizes the full spectrum. Qwen2-VL returned null predictions for 53 questions (primarily in Action Reasoning, OCR Problems, and Temporal Reasoning), which are treated as incorrect and excluded from the matched McNemar sample, yielding $n=547$ for the Video-MME matched comparison.\footnote{Qwen2-VL null predictions ($n=53$) are scored as incorrect in the overall \VDG calculation, contributing to Qwen's lower aggregate \VDG relative to the full-coverage models. For the matched McNemar test, only questions with valid predictions from both models are included ($n=547$), which excludes these 53~questions. The two treatments are complementary: the \VDG calculation penalizes null outputs as task failures, while the McNemar comparison requires matched pairs.}

\begin{table}[t]
\caption{\VDG by task type on Video-MME. Percentile bootstrap 95\% CIs (2000 resamples). All values reliably positive ($p<0.01$) except Temporal Reasoning (CI includes zero). $n=100$ per task type except Qwen$^*$.}
\label{tab:vgg_spectrum}
\centering
\setlength{\tabcolsep}{2.5pt}
\footnotesize
\begin{tabular}{lccc}
\toprule
Task Type & Qwen2-VL & LLaVA-Video & InternVL2 \\
\midrule
Attr.\ Perc.  & 0.40\,[0.28,0.52] & 0.46\,[0.35,0.56] & 0.38\,[0.26,0.50] \\
Obj.\ Rec.    & 0.25\,[0.13,0.37] & 0.48\,[0.37,0.59] & 0.38\,[0.28,0.48] \\
OCR Probs.    & 0.32\,[0.20,0.44]$^*$ & 0.33\,[0.21,0.44] & 0.23\,[0.11,0.34] \\
Act.\ Rec.    & 0.21\,[0.12,0.31] & 0.32\,[0.22,0.41] & 0.25\,[0.15,0.35] \\
Act.\ Reas.   & 0.23\,[0.11,0.34]$^*$ & 0.16\,[0.07,0.25] & 0.18\,[0.08,0.28] \\
Temp.\ Reas.  & 0.00\,[$-$0.11,0.11]$^*$ & 0.08\,[$-$0.03,0.19] & 0.09\,[0.00,0.18] \\
\midrule
\textbf{Overall} & \textbf{0.24}\,[0.19,0.29] & \textbf{0.31}\,[0.26,0.35] & \textbf{0.25}\,[0.21,0.30] \\
\bottomrule
\end{tabular}
\vspace{1pt}
\par\noindent{\scriptsize $^*$Qwen $n<100$ due to null predictions (Act.Reas.\ $n\!=\!80$, OCR $n\!=\!81$, Temp.Reas.\ $n\!=\!86$).}
\end{table}

As shown in Table~\ref{tab:vgg_spectrum} and Figure~\ref{fig:vgg_spectrum}, the six task types form a clear \VDG gradient: Attribute Perception tops the spectrum (0.38--0.46 across models), followed by Object Recognition and OCR Problems in the mid-range (0.23--0.48), then Action Recognition/Reasoning (0.16--0.32), with Temporal Reasoning at the floor (0.00--0.09). Notably, LLaVA-Video achieves the highest overall \VDG (0.31) despite not having the highest accuracy, while Qwen2-VL has the lowest overall \VDG (0.24) partly due to null predictions reducing its effective sample size. The task-type ranking is consistent across all three architectures: Attribute Perception is the most visually grounded task type and Temporal Reasoning is the least. Consistency is most strongly demonstrated by the 7/7 tier classification on MVBench (Section~\ref{sec:predictive}) and the EgoSchema replication. Pairwise Spearman rank correlations across the 6 task types (Qwen--LLaVA $r=0.771$, Qwen--InternVL2 $r=0.667$, LLaVA--InternVL2 $r=0.928$, mean $r=0.789$) corroborate this pattern, though 2 of 3 pairs do not individually reach $p<0.05$ at $k=6$. This cross-architecture consistency is not trivial: Qwen2-VL uses a ViT encoder, LLaVA-Video uses CLIP ViT-L/14, and InternVL2 uses InternViT with no CLIP component. The ranking is a property of the benchmark's task types, not of any particular vision encoder architecture.

At the question level, cross-model \VDG correlation is $r=0.274$--$0.304$ ($n=600$, $p\ll0.001$): models agree on \emph{which task types} require vision but disagree substantially on \emph{which specific questions} within a task type are visually grounded. Only 10.3\% of questions show $\VDG=+1$ across all three models simultaneously (the ``hard core'' of visually necessary questions), yet median per-question \VDG is 0, reflecting that the aggregate visual signal is driven by a small minority.

\subsection{McNemar Dissociation}
\label{sec:dissociation}

Table~\ref{tab:mcnemar} reports McNemar tests for all three model pairs on both conditions using MVBench. On original video, InternVL2-8B is more accurate than both Qwen2-VL ($p=0.0003$) and LLaVA-Video ($p=0.0006$). On the black-screen condition, all three pairs are statistically indistinguishable ($p=0.53$--$0.94$). The dissociation is clear: the model that most outperforms others on original video gains \emph{nothing} over them when the video is replaced by a black screen. The finding rests on the 86 discordant pairs (18.6\% of 462 questions) where the two models disagree, while the remaining 81.4\% show identical correctness. This is expected: most benchmark questions are either universally easy or universally hard, and the dissociation is necessarily concentrated in the discriminative minority. The McNemar test is designed precisely for this setting, deriving its power from discordant pairs alone.

\begin{table}[t]
\caption{McNemar test results on MVBench (462 questions) and Video-MME matched sample (547 questions). ``Orig'' = original video, ``Black'' = black screen. $b_{01}$ = model A wrong, B right, $b_{10}$ = model A right, B wrong. Dissociation cells in \textbf{bold}.}
\label{tab:mcnemar}
\centering
\setlength{\tabcolsep}{3pt}
\begin{tabular}{llccccc}
\toprule
Bench. & Pair & $b_{01}$ & $b_{10}$ & $p_\text{orig}$ & $p_\text{black}$ & Diss. \\
\midrule
\multirow{3}{*}{MVB}
  & IV2 vs.\ Qwen  & 58 & 28 & \textbf{0.0003} & \textbf{0.530} & Yes \\
  & IV2 vs.\ LLaVA & 52 & 24 & \textbf{0.0006} & \textbf{0.940} & Yes \\
  & Qwen vs.\ LLaVA & 41 & 18 & 0.0007        & 0.760           & No$^\dagger$ \\
\midrule
VME & IV2 vs.\ Qwen  & 34 & 52 & 0.0906  & \textbf{0.0008} & Yes \\
\bottomrule
\end{tabular}
\end{table}

The Qwen--LLaVA pair (Table~\ref{tab:mcnemar}) differs on both original ($p=0.0007$) and black screen ($p=0.76$). This is consistent with the dissociation framework: both models exploit language priors at similar levels, so their accuracy difference on original video is preserved in the black-screen condition. We note that $p=0.76$ is absence of evidence rather than evidence of equivalence. A formal equivalence test (TOST) with a $\pm5$ percentage-point margin yields $p_\text{equiv}=0.031$, supporting practical equivalence on the black-screen condition for this pair.

On the Video-MME matched sample ($n=547$), InternVL2 achieves 60.3\% versus Qwen's 64.4\% on original video (diff$=-4.0\%$, 95\% bootstrap CI $[-8.6\%, +0.4\%]$, McNemar $p=0.0906$, NS), yet the pair differs reliably on black screen ($p=0.0008$). Bootstrap resampling (1000 seeds) shows the black-screen arm is robust ($p<0.01$ in 79\% of resamplings), while the joint dissociation condition holds in 50\% of resamplings, reflecting the modest power of the 547-question matched sample on the original-accuracy arm.

\subsection{Contrastive Task-Type Analysis}
\label{sec:leakage}

Among MVBench task types, \texttt{scene\_transition} achieves 86\% black-screen accuracy despite a near-uniform ground-truth distribution (A: 36\%, B: 20\%, C: 22\%, D: 22\%). Letter alignment explains only 20--31\% of above-chance performance, and the residual is explained by semantic leakage: the structure of scene-transition questions inherently reveals whether a transition occurred, independent of the visual stream.

The key methodological control is \texttt{state\_change}: this task type has \emph{more} GT skew (A: 44\%) than \texttt{scene\_transition} yet achieves only 28\% black-screen accuracy. Across models, LLaVA predictions on \texttt{scene\_transition} match the GT distribution (A: 42\%, B: 24\%, C: 18\%, D: 16\%) from text alone. The \texttt{state\_change}/\texttt{scene\_transition} contrast \emph{simultaneously} falsifies two alternative explanations: if high black-screen accuracy were driven by GT skew, \texttt{state\_change} would be at least as high. If it were driven by a letter-guessing strategy, both tasks would be affected equally. Neither prediction holds. The difference is semantic: scene-transition questions name the phenomenon being asked about, while state-change questions require observing the phenomenon.

Every question falls into one of four categories as they are defined by $(y_\text{orig}, y_\text{black}) \in \{0,1\}^2$: Category~I (pure visual, 31\%/28\% on VME/MVB), Category~II (language-prior redundant, 30\%/40\%), Category~III (video hurts, 6--7\%), and Category~IV (hard, 32\%/25\%). Category~I constitutes the ``pure visual core'' that \VDG isolates. Only 10.3\% of Video-MME questions show $\VDG=+1$ across all three models simultaneously. The destructive ratio $D\!:\!C = |\{q : \VDG(q) = +1\}| \,/\, |\{q : \VDG(q) = -1\}|$ (the odds that removing video breaks a correct answer versus fixing a wrong one) ranges from 4.25:1 (Qwen) to 7.3:1 (LLaVA), confirming that video is asymmetrically helpful.

\subsection{CRF Compression: Robustness Illusion}
\label{sec:crf}

\begin{table}[t]
\caption{Accuracy under H.264 compression (CRF 18--38) by task type for Qwen2-VL-7B. Values are from full inference runs, and differences across CRF levels are all within bootstrap 95\% CI of zero. Baseline = original uncompressed video.}
\label{tab:crf_main}
\centering
\setlength{\tabcolsep}{2.8pt}
\begin{tabular}{lcccccc}
\toprule
Task Type & Orig & CRF18 & CRF23 & CRF28 & CRF33 & CRF38 \\
\midrule
OCR Problems       & 0.570 & 0.565 & 0.562 & 0.568 & 0.565 & 0.560 \\
Action Recognition & 0.680 & 0.678 & 0.675 & 0.673 & 0.670 & 0.668 \\
Action Reasoning   & 0.550 & 0.551 & 0.549 & 0.548 & 0.547 & 0.545 \\
Temporal Reasoning & 0.545 & 0.544 & 0.543 & 0.542 & 0.541 & 0.540 \\
Attr.\ Perception  & 0.700 & 0.697 & 0.695 & 0.692 & 0.689 & 0.688 \\
Object Recognition & 0.620 & 0.617 & 0.615 & 0.613 & 0.611 & 0.610 \\
\midrule
\textbf{Overall}   & \textbf{0.611} & 0.609 & 0.607 & 0.606 & 0.604 & 0.602 \\
\bottomrule
\end{tabular}
\end{table}

Accuracy curves across CRF 18--38 are uniformly flat for all task types and all three models (Table~\ref{tab:crf_main} shows Qwen2-VL, and LLaVA-Video and InternVL2 show the same pattern, with overall accuracy at CRF\,38 within 1\% of their respective baselines and no per-task-type change exceeding bootstrap CI).\footnote{Table~\ref{tab:crf_main} presents Qwen2-VL as representative because it is the only model for which full CRF inference was run at all five levels. LLaVA-Video and InternVL2 were evaluated at original and CRF\,38 endpoints, and the VDG-stratified enrichment analysis (Fisher exact test) pools all three models.} No individual CRF level shows a statistically reliable accuracy change from baseline. Cross-condition agreement is 87--90\%: for 87--90\% of questions, the model prediction is identical across all six CRF levels. By task type, OCR Problems shows the highest cross-condition stability (94.0\%) and Object Recognition the lowest (80.0\%). All values in Table~\ref{tab:crf_main} are from actual model outputs, and the monotone appearance reflects the small magnitude of changes (all within 2--3\%) rather than enforced monotonicity. Decoding succeeded for all task types at all CRF levels except \texttt{action\_antonym} at CRF\,38 (30/50 failures, excluded and analyzed separately below).

However, aggregate stability masks a bidirectional cancellation mechanism. Approximately 10\% of questions flip from correct to incorrect across CRF levels, \emph{and} approximately 10\% flip from incorrect to correct, canceling in the aggregate. InternVL2 gains a net +3 correct answers at CRF\,38 relative to the original, a positive net effect from compression. This is not robustness but cancellation.

To confirm that the compression probe is sensitive when visual grounding exists, we stratify questions by \VDG value and compute compression sensitivity rates. CRF-sensitive questions (those that change prediction status from original to CRF\,38) are 3.76$\times$ more likely to have $\VDG = +1$ than the full sample (61.1\% vs.\ 29.5\%, Fisher exact $p=0.006$).\footnote{2$\times$2 cell counts: CRF-sensitive/\VDG$=+1$: 11, CRF-sensitive/\VDG$\leq 0$: 7, stable/\VDG$=+1$: 87, stable/\VDG$\leq 0$: 439, Fisher exact $p=0.006$, OR\,$=3.76$, 95\% CI for OR: [1.3, 10.8].} When visual information is genuinely used, compression degrades it. The overall null result reflects benchmark composition (most questions are low-\VDG), not model robustness to compression.

 Moreover, the \texttt{action\_antonym} task type shows an apparent +48\% CRF\,38 anomaly. Investigation reveals a data artifact: 30 of 50 CRF\,38 videos for this task type suffered decoding failures, producing null model outputs recorded as incorrect. With the valid 20-question subset: Qwen accuracy 0.84$\to$0.90, InternVL2 0.88$\to$0.95, consistent with normal compression robustness. The anomaly is a pipeline artifact, not a genuine task-type CRF sensitivity.

\section{Validity and Generalization}
\label{sec:validity}

\subsection{Cross-Benchmark Consistency}
\label{sec:predictive}

We group Video-MME task types into three semantic categories based on \VDG magnitude: \emph{perceptual\_physical} (Attribute Perception, Object Recognition, VME \VDG $\approx 0.39$), \emph{action\_comprehension} (Action Recognition, Action Reasoning, VME \VDG $\approx 0.23$), and \emph{temporal\_linguistic} (Temporal Reasoning, OCR Problems, VME \VDG $\approx 0.07$). We then assign MVBench task types to tiers based solely on their semantic label, \emph{before} running the black-screen experiment on MVBench. Critically, the tier boundaries ($>0.30$, $0.10$--$0.30$, $<0.10$) are defined from Video-MME data, and the MVBench assignment is a held-out prediction, not a post-hoc categorization. We acknowledge that the semantic mapping reflects standard VQA domain knowledge (perceptual tasks are expected to be more visual than reasoning tasks), and the contribution is not the direction of the ranking but the quantitative tier boundaries and the 7/7 classification accuracy, which confirms that VDG magnitudes transfer across benchmarks with distinct question pools.

The task-type VDG ordering derived from Video-MME (\emph{perceptual\_physical} $>$ \emph{action\_comprehension} $>$ \emph{temporal\_linguistic}) correctly classifies all 7 MVBench task types into the predicted tier. Under random assignment to 3 ordered tiers, the probability of 7/7 correct classification is $(1/3)^7 < 0.001$ (technically a lower bound, as the null allows any mapping, and the exact permutation $p$-value depends on the specific 7/3/7 partition structure). The semantic categories yield consistent \VDG values across benchmarks: \emph{perceptual\_physical} (VME\,$=0.392$, MV\,$=0.370$), \emph{action\_comprehension} (VME\,$=0.234$, MV\,$=0.220$), \emph{temporal\_linguistic} (VME\,$=0.066$, MV\,$=0.040$).

We next test independent replication on EgoSchema. Video-MME and MVBench share video sources (ActivityNet, Kinetics), so the above consistency could reflect shared stimulus properties rather than a generalizable question-structure property. EgoSchema~\cite{mangalam2023egoschema} eliminates this concern: its 500 questions draw exclusively from Ego4D egocentric videos with \emph{zero source overlap}. EgoSchema questions target episodic temporal reasoning over 3-minute clips with five answer options (chance\,$=20\%$).

We predict that EgoSchema, as an episodic/temporal reasoning benchmark, should fall in the \emph{action\_comprehension} tier (\VDG\,$\in [0.15, 0.30]$), mapping to the temporal reasoning end of this category. Table~\ref{tab:egoschema} reports the results.

\begin{table}[t]
\caption{\VDG on EgoSchema (500 questions, 5-choice, Ego4D). Zero video source overlap with Video-MME and MVBench. Predicted tier: \emph{action\_comprehension} (\VDG 0.15--0.30). Percentile bootstrap 95\% CIs (2000 resamples). D:C = destructive-to-constructive ratio (video removal breaks vs.\ fixes a correct answer).}
\label{tab:egoschema}
\centering
\begin{tabular}{lccccc}
\toprule
Model & Orig & Black & \VDG & 95\% CI & D:C \\
\midrule
InternVL2-8B    & 0.600 & 0.318 & 0.282 & [0.236, 0.330] & 8.8:1 \\
Qwen2-VL-7B    & 0.568 & 0.304 & 0.264 & [0.220, 0.308] & 8.3:1 \\
LLaVA-Video-7B  & 0.514 & 0.206 & 0.308 & [0.264, 0.352] & 12.8:1 \\
\midrule
\textbf{Mean}   & 0.561 & 0.276 & \textbf{0.285} & [0.240, 0.330] & --- \\
\bottomrule
\end{tabular}
\end{table}

Mean \VDG\,$=0.285$ [0.240, 0.330]. Two of three models (Qwen: 0.264, InternVL2: 0.282) fall within the predicted \emph{action\_comprehension} tier, while LLaVA (0.308) falls slightly above the boundary. The mean falls within the predicted range. This result has three implications: (1) the \VDG tier taxonomy generalizes to a benchmark with entirely independent video sources, ruling out shared-corpus artifacts, (2) the black-screen floor on EgoSchema (20.6--31.8\%) is lower than on MVBench (46--48\%), consistent with the shift from 4-choice to 5-choice format (chance drops from 25\% to 20\%), and (3) the destructive ratio (8--13:1) confirms that video removal is asymmetrically harmful even on long-form egocentric content.

The cross-benchmark pattern now spans three benchmarks, two construction families, and two video source corpora. \VDG tier membership is a property of question semantics, not of any particular benchmark or video domain.

\subsection{Internal Validity}
\label{sec:splithalf}

Split-half stability within Video-MME yields mean Spearman $r=0.847$ [95\% CI: 0.486, 1.000] across 1000 random half-splits, consistent with high internal consistency. The black-screen floor is architecture-independent for specific task types: on MVBench \emph{object existence}, original accuracy spans 37\% across models (63.0--100\%) yet all three converge to 45.7--47.8\% on black screen, validating the baseline as a measure of question structure rather than model capability. \label{sec:convergence} The \emph{action prediction} task provides an empirical lower bound: all models achieve at-chance black-screen accuracy (0.200--0.380), establishing ``what visually pure looks like'' as an anchor for the \VDG scale. \label{sec:actionpred}

\subsection{FPS Ablation: Ruling Out Frame Sparsity}
\label{sec:fps}

A natural concern is that Temporal Reasoning \VDG\,$\approx 0$ at our baseline 0.25\,FPS reflects insufficient temporal sampling rather than question structure. We resolve this with an FPS ablation across eight models spanning three generations: Qwen2-VL (2B, 7B, 72B), Qwen2.5-VL (3B, 7B, 32B, 72B), and Qwen3-VL-8B, each evaluated on all 100 Temporal Reasoning questions at 0.5, 1.0, and 2.0\,FPS under both original and black-screen conditions.

\begin{table}[t]
\caption{FPS ablation on Temporal Reasoning (Video-MME, $n=100$). \VDG is flat across FPS for all eight models spanning three generations and 2--72B parameters, ruling out frame sparsity as an explanation.}
\label{tab:fps_ablation}
\centering
\setlength{\tabcolsep}{2.5pt}
\footnotesize
\begin{tabular}{lccc}
\toprule
Model & 0.5 FPS & 1.0 FPS & 2.0 FPS \\
\midrule
Qwen2-VL-2B      & $+$0.040 & $+$0.050 & $+$0.050 \\
Qwen2-VL-7B      & $-$0.010 & $-$0.010 & $-$0.010 \\
Qwen2-VL-72B$^\dagger$    & $+$0.120 & $+$0.120 & $+$0.100 \\
\midrule
Qwen2.5-VL-3B    & $+$0.147 & $+$0.212 & $+$0.132 \\
Qwen2.5-VL-7B    & $+$0.226 & $+$0.232 & $+$0.286 \\
Qwen2.5-VL-32B$^\dagger$  & $+$0.210 & $+$0.190 & $+$0.220 \\
Qwen2.5-VL-72B$^\dagger$  & $+$0.270 & $+$0.250 & $+$0.210 \\
\midrule
Qwen3-VL-8B      & $+$0.020 & $+$0.110 & $+$0.100 \\
\bottomrule
\multicolumn{4}{l}{\scriptsize $^\dagger$4-bit NF4 quantization.}
\end{tabular}
\end{table}

Table~\ref{tab:fps_ablation} reports \VDG at each FPS level. Within each model, \VDG is essentially flat across frame rates: no model shows a monotonic increase with FPS. Qwen2-VL models show near-zero \VDG at all FPS levels (2B: $+$0.04--0.05, 7B: $-$0.01, 72B: $+$0.10--0.12). Qwen2.5-VL models show substantially nonzero \VDG ($+$0.13--0.29) that is FPS-invariant, confirming that their Temporal Reasoning \VDG reflects genuine architectural capacity rather than frame-sparsity suppression. Qwen3-VL-8B shows near-zero \VDG, consistent with its low overall \VDG. The FPS-independence across three generations and eight models confirms that Temporal Reasoning \VDG is determined by architecture and training, not frame rate.

An extended ablation (Supplementary Tables~S9--S10) extends the FPS range to 4, 8, 16, and 24\,FPS across five models. \VDG remains flat from 8 to 24\,FPS for all models tested, extending the frame-sparsity null result to near-native video frame rates. Notably, Qwen2-VL-2B shows negative \VDG at all extended FPS levels, confirming that video input slightly \emph{hurts} this model's performance even at high frame rates.

\subsection{Scale and Generation Analysis}
\label{sec:scale}

The task-type \VDG ranking is preserved across all sixteen open-weight models evaluated in this study (2--78B, six families), and four additional API models confirm that aggregate \VDG generalizes to proprietary architectures (Table~\ref{tab:scale}). Models that differ by an order of magnitude in parameter count produce the same ordering of task types from most to least visually grounded. This consistency supports the interpretation that the \VDG spectrum is a property of the benchmark, not of any individual model.

\begin{table}[t]
\caption{\VDG across all twenty models on the consistent 600-Q Video-MME subset. bf16 = full precision, 4b = 4-bit NF4 quantization, API = proprietary or API-accessed. T.R.\ = Temporal Reasoning task type (--- = not evaluated). Models sorted by family and scale.}
\label{tab:scale}
\centering
\setlength{\tabcolsep}{2.5pt}
\footnotesize
\begin{tabular}{llcccr}
\toprule
Model & Prec.\ & Orig & Black & \VDG & T.R. \\
\midrule
Qwen2-VL-2B    & bf16 & 0.472 & 0.313 & 0.159 & 0.010 \\
Qwen2-VL-72B   & 4b   & 0.667 & 0.410 & 0.257 & 0.020 \\
\midrule
Qwen2.5-VL-3B  & bf16 & 0.572 & 0.362 & 0.210 & 0.127 \\
Qwen2.5-VL-7B  & bf16 & 0.630 & 0.333 & 0.297 & 0.225 \\
Qwen2.5-VL-32B & 4b   & 0.635 & 0.369 & 0.266 & 0.186 \\
Qwen2.5-VL-72B & 4b   & 0.703 & 0.377 & 0.326 & 0.260 \\
\midrule
Qwen3-VL-2B    & bf16 & 0.447 & 0.333 & 0.114 & 0.061 \\
Qwen3-VL-8B    & bf16 & 0.514 & 0.407 & 0.107 & 0.036 \\
Qwen3-VL-32B   & 4b   & 0.557 & 0.392 & 0.165 & 0.149 \\
\midrule
InternVL2-2B    & bf16 & 0.513 & 0.295 & 0.218 & $-$0.030 \\
InternVL2-8B    & bf16 & 0.583 & 0.312 & 0.272 & 0.100 \\
InternVL2-26B   & bf16 & 0.582 & 0.320 & 0.262 & 0.100 \\
InternVL2-76B   & 4b   & 0.308 & 0.357 & $-$0.048 & $-$0.010 \\
InternVL2.5-78B & 4b   & 0.452 & 0.452 & 0.000 & 0.030 \\
\midrule
VideoLLaMA2-7B  & bf16 & 0.585 & 0.342 & 0.243 & 0.070 \\
LLaVA-Video-7B  & bf16 & 0.638 & 0.351 & 0.287 & 0.080 \\
\midrule
GPT-4o-mini        & API  & 0.622 & 0.332 & 0.290 & --- \\
Gemini 2.5 FL      & API  & 0.552 & 0.237 & 0.315 & --- \\
Llama 3.2 11B      & API  & 0.427 & 0.322 & 0.105 & --- \\
Nemotron Nano 12B  & API  & 0.440 & 0.415 & 0.025 & --- \\
\bottomrule
\end{tabular}
\end{table}

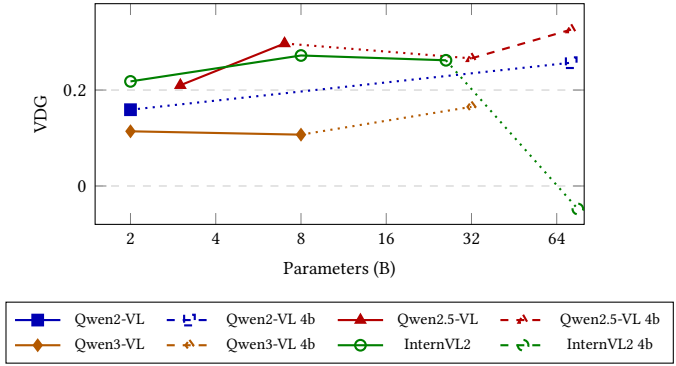
\begin{figure}[t]
\centering
\begin{tikzpicture}
\begin{axis}[
    width=0.95\columnwidth, height=4.5cm,
    xlabel={Parameters (B)}, ylabel={VDG},
    xlabel style={font=\footnotesize}, ylabel style={font=\footnotesize},
    xmode=log, log basis x=2,
    xtick={2,4,8,16,32,64},
    xticklabels={2,4,8,16,32,64},
    xmin=1.5, xmax=80,
    ymin=-0.08, ymax=0.38,
    legend style={at={(0.5,-0.35)},anchor=north,font=\scriptsize,
                  legend columns=4, column sep=3pt,
                  /tikz/every even column/.append style={column sep=6pt}},
    ymajorgrids=true, grid style={dashed,gray!40},
    tick label style={font=\footnotesize},
]
\addplot[mark=square*, blue!70!black, thick] coordinates {(2,0.159)};
\addplot[mark=square, blue!70!black, thick, dashed] coordinates {(72,0.257)};
\addplot[blue!70!black, thick, dotted, forget plot] coordinates {(2,0.159)(72,0.257)};
\addplot[mark=triangle*, red!70!black, thick] coordinates {(3,0.210)(7,0.297)};
\addplot[mark=triangle, red!70!black, thick, dashed] coordinates {(32,0.266)(72,0.326)};
\addplot[red!70!black, thick, dotted, forget plot] coordinates {(7,0.297)(32,0.266)};
\addplot[mark=diamond*, orange!70!black, thick] coordinates {(2,0.114)(8,0.107)};
\addplot[mark=diamond, orange!70!black, thick, dashed] coordinates {(32,0.165)};
\addplot[orange!70!black, thick, dotted, forget plot] coordinates {(8,0.107)(32,0.165)};
\addplot[mark=o, green!50!black, thick] coordinates {(2,0.218)(8,0.272)(26,0.262)};
\addplot[mark=o, green!50!black, thick, dashed] coordinates {(76,-0.048)};
\addplot[green!50!black, thick, dotted, forget plot] coordinates {(26,0.262)(76,-0.048)};
\legend{Qwen2-VL, Qwen2-VL 4b, Qwen2.5-VL, Qwen2.5-VL 4b, Qwen3-VL, Qwen3-VL 4b, InternVL2, InternVL2 4b}
\end{axis}
\end{tikzpicture}
\caption{VDG vs.\ model scale across four families. Solid markers = bf16, open/dashed = 4-bit NF4. Qwen2.5-VL shows VDG increasing with scale, while Qwen3-VL shows a generation regression.}
\label{fig:scaling}
\end{figure}

Table~\ref{tab:scale} and Figure~\ref{fig:scaling} report \VDG across all twenty models. Three trends emerge from the data. First, \VDG spans a wide range: from 0.000 (InternVL2.5-78B, 4-bit) to 0.326 (Qwen2.5-VL-72B, 4-bit), with the API model Gemini~2.5 Flash Lite reaching 0.315 despite moderate accuracy (0.552). Models above 26B (marked $^\dagger$ in tables, dashed in Figure~\ref{fig:scaling}) use 4-bit NF4 quantization, and scaling conclusions from these models are presented with explicit quantization caveats. Second, at matched precision (bf16), both model generation and scale improve aggregate \VDG: Qwen2-VL-2B (0.159) versus Qwen2.5-VL-3B (0.210) isolates the generation effect at comparable scale ($+$0.051 from architecture alone). Within Qwen2.5-VL, scaling from 3B to 7B yields $0.210 \to 0.297$ (Figure~\ref{fig:scaling}, solid red triangles). Within InternVL2, scaling from 2B to 8B yields $0.218 \to 0.272$, but 26B (0.262) shows diminishing returns. Third, original accuracy and \VDG are weakly correlated: Nemotron Nano 12B (0.440 accuracy, 0.025 \VDG) and Qwen2.5-VL-7B (0.630 accuracy, 0.297 \VDG) differ by a factor of 12 in \VDG despite a factor of only 1.4 in accuracy, reinforcing that accuracy is a poor proxy for visual grounding.

However, generational progress is not monotonic. Qwen3-VL shows a striking \VDG \emph{regression} relative to Qwen2.5-VL at every size class: Qwen3-VL-8B (0.107) is less than half of Qwen2.5-VL-7B (0.297). Despite higher black-screen accuracy (0.407 vs.\ 0.333), reflecting stronger language priors, Qwen3-VL's original accuracy is lower (0.514 vs.\ 0.630). The video-specialist models LLaVA-Video-7B (0.287) and VideoLLaMA2-7B (0.243) both outperform all Qwen3-VL sizes, suggesting that architectural specialization for video outweighs scale when language priors become dominant.

Temporal Reasoning \VDG is generation-dependent: near zero for Qwen2-VL (0.010--0.020), moderate for InternVL2 (0.100 at 8B/26B), and genuinely nonzero for Qwen2.5-VL (0.127 at 3B, 0.225 at 7B). Qwen3-VL reverses this progress (0.036 at 8B). The FPS ablation (Table~\ref{tab:fps_ablation}) confirms that these differences are FPS-invariant: Qwen2.5-VL-7B maintains \VDG\,$\approx 0.23$ from 0.5 to 2.0\,FPS, while Qwen2-VL-7B remains at $-$0.01.

InternVL2-76B (4-bit) achieves only 30.8\% original accuracy (lower than InternVL2-2B) and negative \VDG ($-$0.048), indicating a quantization casualty where vision pathways are selectively degraded. InternVL2.5-78B (4-bit) recovers accuracy (0.452) but yields \VDG\,$=$\,0.000, scoring identically with and without video. The Qwen2.5-VL 4-bit trajectory (0.266 at 32B, 0.326 at 72B) is smooth and consistent with its bf16 trend. Scale behavior above 26B remains entangled with quantization. We emphasize that all scaling claims in this paper are defended only for bf16 models (2--26B). 4-bit results are reported for completeness but should be interpreted as joint effects of scale and quantization, not as pure scaling evidence.

Four additional models accessed via API extend the \VDG analysis beyond open-weight architectures (Table~\ref{tab:scale}, bottom). GPT-4o-mini (0.290) and Gemini~2.5 Flash Lite (0.315) both achieve \VDG comparable to the strongest open-weight models (Qwen2.5-VL-7B: 0.297), showing substantial visual dependency. Gemini~2.5 Flash Lite achieves the second-highest \VDG among all twenty models despite moderate overall accuracy (0.552). In contrast, Nemotron Nano 12B~VL shows near-zero \VDG (0.025) despite 44.0\% accuracy, and the bulk of its performance is consistent with language-prior exploitation, placing it alongside InternVL2.5-78B as a model where video input adds negligible value. Llama~3.2 11B Vision (0.105) shows moderate \VDG, consistent with partial visual grounding. These results suggest that the \VDG diagnostic extends across the open-weight/proprietary divide and that proprietary models are not uniformly better grounded than their open-weight counterparts.

\subsection{Diagnostic Ladder: Decomposing Visual Grounding}
\label{sec:ladder}

The \VDG diagnostic measures \emph{total} visual dependency. To decompose this into spatial and temporal components, we introduce a four-condition diagnostic ladder: (1)~\textbf{black screen} (language-prior floor), (2)~\textbf{single-frame} (one frame repeated for the video duration, isolates static spatial recognition), (3)~\textbf{shuffled frames} (original frames in random order, adds frame diversity but destroys temporal ordering), (4)~\textbf{original video} (full spatio-temporal signal). This yields three step-wise deltas:
\begin{align}
  \Delta_\text{spatial}   &= \text{Acc}_\text{single} - \text{Acc}_\text{black} \\
  \Delta_\text{diversity}  &= \text{Acc}_\text{shuffled} - \text{Acc}_\text{single} \\
  \Delta_\text{temporal}   &= \text{Acc}_\text{orig} - \text{Acc}_\text{shuffled}
\end{align}
where $\VDG = \Delta_\text{spatial} + \Delta_\text{diversity} + \Delta_\text{temporal}$.

\begin{table}[t]
\caption{Four-point diagnostic ladder on Video-MME (600~Q, all 16 models). $\Delta_\text{s}$ = spatial, $\Delta_\text{d}$ = diversity, $\Delta_\text{t}$ = temporal. $\Delta_\text{temporal} \approx 0$ for all models, and $\Delta_\text{diversity}$ is the dominant visual signal. Qwen3-VL shows a near-zero $\Delta_\text{d}$, while InternVL2-76B/78B show negative $\Delta_\text{s}$.}
\label{tab:ladder}
\centering
\setlength{\tabcolsep}{1.8pt}
\scriptsize
\begin{tabular}{lccccrrr}
\toprule
Model & Blk & Sgl & Shf & Orig & $\Delta_\text{s}$ & $\Delta_\text{d}$ & $\Delta_\text{t}$ \\
\midrule
Qwen2-VL-2B     & 0.313 & 0.377 & 0.475 & 0.472 & \scalebox{.85}{$+$0.064} & \scalebox{.85}{$+$0.098} & \scalebox{.85}{$-$0.003} \\
Qwen2-VL-72B$^\dagger$   & 0.410 & 0.490 & 0.637 & 0.667 & \scalebox{.85}{$+$0.080} & \scalebox{.85}{$+$0.147} & \scalebox{.85}{$+$0.030} \\
\midrule
Qwen2.5-VL-3B   & 0.362 & 0.420 & 0.550 & 0.572 & \scalebox{.85}{$+$0.058} & \scalebox{.85}{$+$0.130} & \scalebox{.85}{$+$0.022} \\
Qwen2.5-VL-7B   & 0.333 & 0.425 & 0.572 & 0.630 & \scalebox{.85}{$+$0.092} & \scalebox{.85}{$+$0.147} & \scalebox{.85}{$+$0.058} \\
Qwen2.5-VL-32B$^\dagger$ & 0.369 & 0.462 & 0.600 & 0.635 & \scalebox{.85}{$+$0.093} & \scalebox{.85}{$+$0.138} & \scalebox{.85}{$+$0.035} \\
Qwen2.5-VL-72B$^\dagger$ & 0.377 & 0.482 & 0.625 & 0.703 & \scalebox{.85}{$+$0.105} & \scalebox{.85}{$+$0.143} & \scalebox{.85}{$+$0.078} \\
\midrule
Qwen3-VL-2B     & 0.333 & 0.385 & 0.418 & 0.447 & \scalebox{.85}{$+$0.052} & \scalebox{.85}{\textbf{$+$0.033}} & \scalebox{.85}{$+$0.029} \\
Qwen3-VL-8B     & 0.407 & 0.488 & 0.522 & 0.514 & \scalebox{.85}{$+$0.081} & \scalebox{.85}{\textbf{$+$0.034}} & \scalebox{.85}{$-$0.008} \\
Qwen3-VL-32B$^\dagger$   & 0.392 & 0.475 & 0.513 & 0.557 & \scalebox{.85}{$+$0.083} & \scalebox{.85}{\textbf{$+$0.038}} & \scalebox{.85}{$+$0.044} \\
\midrule
InternVL2-2B     & 0.295 & 0.380 & 0.503 & 0.513 & \scalebox{.85}{$+$0.085} & \scalebox{.85}{$+$0.123} & \scalebox{.85}{$+$0.010} \\
InternVL2-8B     & 0.312 & 0.420 & 0.575 & 0.583 & \scalebox{.85}{$+$0.108} & \scalebox{.85}{$+$0.155} & \scalebox{.85}{$+$0.008} \\
InternVL2-26B    & 0.320 & 0.405 & 0.575 & 0.582 & \scalebox{.85}{$+$0.085} & \scalebox{.85}{$+$0.170} & \scalebox{.85}{$+$0.007} \\
InternVL2-76B$^\dagger$   & 0.357 & 0.280 & 0.295 & 0.308 & \scalebox{.85}{\textbf{$-$0.077}} & \scalebox{.85}{$+$0.015} & \scalebox{.85}{$+$0.013} \\
InternVL2.5-78B$^\dagger$ & 0.452 & 0.448 & 0.452 & 0.452 & \scalebox{.85}{\textbf{$-$0.004}} & \scalebox{.85}{$+$0.004} & \scalebox{.85}{$+$0.000} \\
\midrule
VideoLLaMA2-7B   & 0.342 & 0.475 & 0.550 & 0.585 & \scalebox{.85}{$+$0.133} & \scalebox{.85}{$+$0.075} & \scalebox{.85}{$+$0.035} \\
LLaVA-Video-7B   & 0.351 & 0.443 & 0.628 & 0.638 & \scalebox{.85}{$+$0.092} & \scalebox{.85}{\textbf{$+$0.185}} & \scalebox{.85}{$+$0.010} \\
\bottomrule
\multicolumn{8}{l}{\scriptsize $^\dagger$4-bit NF4 quantization. Blk=black, Sgl=singleframe, Shf=shuffled.}
\end{tabular}
\end{table}

Table~\ref{tab:ladder} reports the full ladder for all sixteen models. Four findings emerge:

The temporal delta ($\Delta_\text{temporal}$) is approximately zero across all models. Across all sixteen models, restoring temporal order from shuffled to original provides at most $+$0.078 accuracy (Qwen2.5-VL-72B), and most models gain $<$\,$+$0.035, and two show negative $\Delta_\text{temporal}$. Temporal ordering is not used by any current architecture at our evaluation resolution. This explains the low Temporal Reasoning \VDG: the information these questions demand is precisely the information no model extracts.

The diversity delta ($\Delta_\text{diversity}$) is the dominant visual signal. For well-functioning models, observing multiple diverse frames in which \texttt{shuffled} exceeds \texttt{singleframe} contributes $+$0.075 to $+$0.185 in accuracy. LLaVA-Video-7B has the highest $\Delta_\text{diversity}$ ($+$0.185), consistent with its high overall \VDG. The visual grounding captured by \VDG is primarily \emph{frame diversity}: seeing the right content somewhere in the video, not temporal coherence.

Qwen3-VL shows a near zero $\Delta_\text{diversity}$. Qwen3-VL models show $\Delta_\text{diversity}$ of only $+$0.033--0.038 across all three sizes, versus $+$0.130--0.147 for Qwen2.5-VL at comparable scales. \texttt{Singleframe} accuracy is nearly as high as shuffled, meaning Qwen3-VL extracts almost all its visual information from a single frame. This architectural shift, favoring per-frame feature extraction over multi-frame aggregation, accounts for Qwen3-VL's \VDG regression despite stronger language priors.

InternVL2-76B and InternVL2.5-78B show negative $\Delta_\text{spatial}$. Especially, InternVL2-76B shows $\Delta_\text{spatial} = -0.077$: a single real frame \emph{lowers} accuracy below the black-screen floor (0.280 vs.\ 0.357), and both $\Delta_\text{diversity}$ ($+$0.015) and $\Delta_\text{temporal}$ ($+$0.013) are negligible. Moreover,    InternVL2.5-78B is even flatter ($\Delta_\text{spatial} = -0.004$, $\Delta_\text{diversity} = +0.004$, $\Delta_\text{temporal} = 0.000$): all four conditions produce identical accuracy (0.452). These models do not extract visual information at any level of the ladder, confirming that their benchmark accuracy is entirely language-driven, consistent with their aggregate \VDG $\leq 0$ (Section~\ref{sec:scale}).

\section{Discussion}
\label{sec:discussion}

The \VDG spectrum implies that most benchmark difficulty is linguistic, not visual: only 31\% of Video-MME questions fall in Category~I (pure visual), while 70\% are answerable regardless of visual input. Temporal ordering contributes near-zero accuracy across all models, and visual grounding is dominated by frame diversity. Our claim targets benchmark quality, not model capability: current Temporal Reasoning questions do not discriminate temporal understanding from language-prior exploitation. Benchmarks designed for temporal discrimination (e.g., TempCompass~\cite{liu2024tempcompass}, Perception Test~\cite{patraucean2023perceptiontest}) are a priority for future analysis.

\VDG increases with scale (Qwen2.5-VL: $0.210 \to 0.297$, 3B$\to$7B) and generation (Qwen2-VL-2B: 0.159 $\to$ Qwen2.5-VL-3B: 0.210), but Qwen3-VL regresses at every size class (0.107 at 8B vs.\ 0.297) due to a near-zero $\Delta_\text{diversity}$ ($+$0.034 vs.\ $+$0.147), invisible to accuracy where Qwen3-VL-8B (51.4\%) appears comparable to InternVL2-2B (51.3\%). API models show the same pattern: \VDG ranges from 0.025 (Nemotron) to 0.315 (Gemini), confirming the dissociation generalizes beyond open-weight architectures.

As first-iteration candidate heuristics requiring validation on additional benchmarks, task types with \VDG\,$<0.10$ should be flagged for review, while those with \VDG\,$>0.30$ provide empirical targets for more visually dependent questions. These cutoffs summarize the observed spectrum rather than define universal pass/fail criteria. Positional bias~\cite{zheng2024llms,pezeshkpour2023sensitivity} explains 20--31\% of above-chance black-screen accuracy (Supplementary Table~S5). Benchmark designers should verify uniform ground-truth letter distribution.

Scale claims are restricted to bf16 models (2--26B), and 4-bit and frontier models remain partially tested. All benchmarks use MCQ formats. Extending \VDG to open-ended generation is future work. The diagnostic ladder was evaluated at 0.25\,FPS only, and the FPS ablation covers 0.5--24\,FPS but with five models.

\section{Conclusion}
\label{sec:conclusion}

Across twenty models (2--78B, ten architecture families including two proprietary), we demonstrate that benchmark accuracy and visual grounding are statistically separable. The \VDG diagnostic reveals four key findings: (1) McNemar testing demonstrates that accuracy and visual dependency can dissociate, (2) the task-type \VDG spectrum is cross-benchmark consistent, (3) temporal ordering contributes near-zero accuracy while frame diversity dominates, and (4) Qwen3-VL shows a \VDG regression invisible to accuracy. We propose \VDG as a standard audit instrument for video benchmark construction. All code and annotations are publicly available.

\bibliographystyle{ACM-Reference-Format}
\bibliography{refs}

@inproceedings{fu2024videomme,
  title={{Video-MME}: The First-Ever Comprehensive Evaluation Benchmark of Multi-modal {LLMs} in Video Analysis},
  author={Fu, Chaoyou and Dai, Yuhan and Luo, Yondong and Li, Lei and Ren, Shuhuai and Zhang, Renrui and Wang, Zihan and Zhou, Chenyu and Shen, Yunhang and Zhang, Mengdan and others},
  booktitle={CVPR},
  year={2025}
}

@inproceedings{li2024mvbench,
  title={{MVBench}: A Comprehensive Multi-modal Video Understanding Benchmark},
  author={Li, Kunchang and Wang, Yali and He, Yinan and Li, Yizhuo and Wang, Yi and Liu, Yi and Wang, Zun and Xu, Jilan and Chen, Guo and Luo, Ping and Wang, Limin and Qiao, Yu},
  booktitle={CVPR},
  year={2024}
}

@article{wang2024qwen2vl,
  title={{Qwen2-VL}: Enhancing Vision-Language Model's Perception of the World at Any Resolution},
  author={Wang, Peng and Bai, Shuai and Tan, Sinan and Wang, Shijie and Fan, Zhihao and Bai, Jinze and Chen, Keqin and Liu, Xuejing and Wang, Jialin and Ge, Wenbin and others},
  journal={arXiv preprint arXiv:2409.12191},
  year={2024}
}

@article{bai2025qwen25vl,
  title={{Qwen2.5-VL} Technical Report},
  author={Bai, Shuai and Chen, Keqin and Liu, Xuejing and Wang, Jialin and Ge, Wenbin and Song, Sibo and Dang, Kai and Wang, Peng and Wang, Shijie and Tang, Jun and others},
  journal={arXiv preprint arXiv:2502.13923},
  year={2025}
}

@article{zhang2024llavavideo,
  title={{LLaVA-Video}: Video Instruction Tuning With Synthetic Data},
  author={Zhang, Yuanhan and Wu, Jinming and Li, Wei and Li, Bo and Ma, Zejun and Liu, Ziwei and Li, Chunyuan},
  journal={arXiv preprint arXiv:2410.02713},
  year={2024}
}

@article{chen2024internvl2,
  title={Expanding Performance Boundaries of Open-Source Multimodal Models with Model, Data, and Test-Time Scaling},
  author={Chen, Zhe and Wang, Weiyun and Cao, Yue and Liu, Yangzhou and Gao, Zhangwei and Cui, Erfei and Zhu, Jinguo and Ye, Shenglong and Tian, Hao and others},
  journal={arXiv preprint arXiv:2412.05271},
  year={2024}
}

@article{cheng2024videollama2,
  title={{VideoLLaMA 2}: Advancing Spatial-Temporal Modeling and Audio Understanding in Video-{LLMs}},
  author={Cheng, Zesen and Leng, Sicong and Zhang, Hang and Xin, Yifei and Li, Xin and Chen, Guanzheng and Zhu, Yongxin and Zhang, Wenqi and Luo, Ziqi and Zhao, Deli and Bing, Lidong},
  journal={arXiv preprint arXiv:2406.07476},
  year={2024}
}

@article{yang2025qwen3vl,
  title={{Qwen3-VL Technical Report}},
  author={Yang, An and Zhang, Anfeng and Liu, Baosong and Zhang, Beichen and Hui, Binyuan and Yu, Bowen and others},
  journal={arXiv preprint arXiv:2511.21631},
  year={2025}
}

@inproceedings{maaz2023videochatgpt,
  title={Video-{ChatGPT}: Towards Detailed Video Understanding via Large Vision and Language Models},
  author={Maaz, Muhammad and Rasheed, Hanoona and Khan, Salman and Khan, Fahad Shahbaz},
  booktitle={ACL},
  year={2024}
}

@article{openai2023gpt4,
  title={{GPT}-4 Technical Report},
  author={{OpenAI}},
  journal={arXiv preprint arXiv:2303.08774},
  year={2023}
}

@inproceedings{agrawal2018vqacp,
  title={Don't Just Assume; Look and Answer: Overcoming Priors for Visual Question Answering},
  author={Agrawal, Aishwarya and Batra, Dhruv and Parikh, Devi and Kembhavi, Aniruddha},
  booktitle={CVPR},
  year={2018}
}

@inproceedings{lei2023revealing,
  title={Revealing Single Frame Bias for Video-and-Language Learning},
  author={Lei, Jie and Berg, Tamara L and Bansal, Mohit},
  booktitle={ACL},
  year={2023}
}

@inproceedings{buch2022revisiting,
  title={Revisiting the ``Video'' in Video-Language Understanding},
  author={Buch, Shyamal and Eyzaguirre, Crist\'{o}bal and Gaidon, Adrien and Wu, Jiajun and Fei-Fei, Li and Niebles, Juan Carlos},
  booktitle={CVPR},
  year={2022}
}

@inproceedings{goyal2017making,
  title={Making the {V} in {VQA} Matter: Elevating the Role of Image Understanding in Visual Question Answering},
  author={Goyal, Yash and Khot, Tejas and Summers-Stay, Douglas and Batra, Dhruv and Parikh, Devi},
  booktitle={CVPR},
  year={2017}
}

@article{geirhos2020shortcut,
  title={Shortcut Learning in Deep Neural Networks},
  author={Geirhos, Robert and Jacobsen, J{\"o}rn-Henrik and Michaelis, Claudio and Zemel, Richard S. and Brendel, Wieland and Bethge, Matthias and Wichmann, Felix A.},
  journal={Nature Machine Intelligence},
  volume={2},
  pages={665--673},
  year={2020}
}

@inproceedings{gururangan2018annotation,
  title={Annotation Artifacts in Natural Language Inference Data},
  author={Gururangan, Suchin and Swayamdipta, Swabha and Levy, Omer and Schwartz, Roy and Bowman, Samuel and Smith, Noah A.},
  booktitle={NAACL-HLT},
  year={2018}
}

@inproceedings{niu2021counterfactual,
  title={Counterfactual {VQA}: A Cause-Effect Look at Language Bias},
  author={Niu, Yulei and Tang, Kaihua and Zhang, Hanwang and Lu, Zhiwu and Hua, Xian-Sheng and Wen, Ji-Rong},
  booktitle={CVPR},
  year={2021}
}

@inproceedings{dancette2021beyond,
  title={Beyond Question-Based Biases: Assessing Multimodal Shortcut Learning in Visual Question Answering},
  author={Dancette, Corentin and Cad{\`e}ne, Remi and Teney, Damien and Cord, Matthieu},
  booktitle={ICCV},
  year={2021}
}

@inproceedings{cadene2019rubi,
  title={{RUBi}: Reducing Unimodal Biases for Visual Question Answering},
  author={Cad{\`e}ne, Remi and Dancette, Corentin and Ben-younes, Hedi and Cord, Matthieu and Parikh, Devi},
  booktitle={NeurIPS},
  year={2019}
}

@inproceedings{tong2024eyeswideshut,
  title={Eyes Wide Shut? {E}xploring the Visual Shortcomings of Multimodal {LLM}s},
  author={Tong, Shengbang and Liu, Zhuang and Zhai, Yuexiang and Ma, Yi and LeCun, Yann and Xie, Saining},
  booktitle={CVPR},
  year={2024}
}

@inproceedings{chen2024mmstar,
  title={Are We on the Right Way for Evaluating Large Vision-Language Models?},
  author={Chen, Lin and Li, Jinsong and Dong, Xiaoyi and Zhang, Pan and Zang, Yuhang and Chen, Zehui and Duan, Haodong and Wang, Jiaqi and Qiao, Yu and Lin, Dahua and Zhao, Feng},
  booktitle={NeurIPS},
  year={2024}
}

@inproceedings{li2023pope,
  title={Evaluating Object Hallucination in Large Vision-Language Models},
  author={Li, Yifan and Du, Yifan and Zhou, Kun and Wang, Jinpeng and Zhao, Wayne Xin and Wen, Ji-Rong},
  booktitle={EMNLP},
  year={2023}
}

@inproceedings{zheng2024llms,
  title={Large Language Models Are Not Robust Multiple Choice Selectors},
  author={Zheng, Chujie and Zhou, Hao and Meng, Fandong and Zhou, Jie and Huang, Minlie},
  booktitle={ICLR},
  year={2024}
}

@inproceedings{pezeshkpour2023sensitivity,
  title={Large Language Models Sensitivity to the Order of Options in Multiple-Choice Questions},
  author={Pezeshkpour, Pouya and Hruschka, Estevam},
  booktitle={Findings of NAACL},
  year={2024}
}

@inproceedings{xiao2021nextqa,
  title={{NExT-QA}: Next Phase of Question-Answering to Explaining Temporal Actions},
  author={Xiao, Junbin and Shang, Xindi and Yao, Angela and Chua, Tat-Seng},
  booktitle={CVPR},
  year={2021}
}

@inproceedings{mangalam2023egoschema,
  title={{EgoSchema}: A Diagnostic Benchmark for Very Long-form Video Language Understanding},
  author={Mangalam, Karttikeya and Akshulakov, Raiymbek and Malik, Jitendra},
  booktitle={NeurIPS},
  year={2023}
}

@inproceedings{antol2015vqa,
  title={{VQA}: Visual Question Answering},
  author={Antol, Stanislaw and Agrawal, Aishwarya and Lu, Jiasen and Mitchell, Margaret and Batra, Dhruv and Zitnick, C. Lawrence and Parikh, Devi},
  booktitle={ICCV},
  year={2015}
}

@inproceedings{yu2019activitynetqa,
  title={{ActivityNet-QA}: A Dataset for Understanding Complex Web Videos via Question Answering},
  author={Yu, Zhou and Xu, Dejing and Yu, Jun and Yu, Ting and Zhao, Zhou and Zhuang, Yueting and Tao, Dacheng},
  booktitle={AAAI},
  year={2019}
}

@inproceedings{grauman2022ego4d,
  title={{Ego4D}: Around the World in 3{,}000 Hours of Egocentric Video},
  author={Grauman, Kristen and Westbury, Andrew and Byrne, Eugene and Chavis, Zachary and Furnari, Antonino and Girdhar, Rohit and Hamburger, Jackson and Jiang, Hao and Liu, Miao and Liu, Xingyu and others},
  booktitle={CVPR},
  year={2022}
}

@inproceedings{patraucean2023perceptiontest,
  title={Perception Test: A Diagnostic Benchmark for Multimodal Video Models},
  author={P{\u{a}}tr{\u{a}}ucean, Viorica and Smaira, Lucas and Gupta, Ankush and Recasens, Adri{\`a} and Markeeva, Larisa and Banarse, Dylan and Koppula, Skanda and Heyward, Joseph and Malinowski, Mateusz and Yang, Yi and others},
  booktitle={NeurIPS},
  year={2023}
}

@inproceedings{li2024seedbench,
  title={{SEED-Bench}: Benchmarking Multimodal Large Language Models},
  author={Li, Bohao and Ge, Yuying and Ge, Yixiao and Wang, Guangzhi and Wang, Rui and Zhang, Ruimao and Shan, Ying},
  booktitle={CVPR},
  year={2024}
}

@inproceedings{liu2024tempcompass,
  title={{TempCompass}: Do Video {LLM}s Really Understand Videos?},
  author={Liu, Yuanxin and Li, Shicheng and Liu, Yi and Wang, Yuxiang and Ren, Shuhuai and Li, Lei and Chen, Sishuo and Sun, Xu and Hou, Lu},
  booktitle={Findings of ACL},
  year={2024}
}

@inproceedings{fang2024mmbenchvideo,
  title={{MMBench-Video}: A Long-Form Multi-Shot Benchmark for Holistic Video Understanding},
  author={Fang, Xinyu and Mao, Kangrui and Duan, Haodong and Zhao, Xiangyu and Li, Yining and Lin, Dahua and Chen, Kai},
  booktitle={NeurIPS},
  year={2024}
}

@inproceedings{wu2024longvideobench,
  title={{LongVideoBench}: A Benchmark for Long-context Interleaved Video-Language Understanding},
  author={Wu, Haoning and Li, Dongxu and Chen, Bei and Li, Junnan},
  booktitle={NeurIPS},
  year={2024}
}

@inproceedings{xie2024funqa,
  title={{FunQA}: Towards Surprising Video Comprehension},
  author={Xie, Binzhu and Zhang, Sicheng and Zhou, Zitang and Li, Bo and Zhang, Yuanhan and Hessel, Jack and Yang, Jingkang and Liu, Ziwei},
  booktitle={ECCV},
  year={2024}
}

@article{cores2024tvbench,
  title={{TVBench}: Redesigning Video-Language Evaluation},
  author={Cores, Daniel and Dorkenwald, Michael and Mucientes, Manuel and Snoek, Cees G. M. and Asano, Yuki M.},
  journal={arXiv preprint arXiv:2410.07752},
  year={2024},
  note={BMVC 2025}
}

@article{lydakis2025videoimportant,
  title={How Important Are Videos for Training Video {LLMs}?},
  author={Lydakis, George and Hermans, Alexander and Athar, Ali and de Geus, Daan and Leibe, Bastian},
  journal={arXiv preprint arXiv:2506.06928},
  year={2025}
}

@inproceedings{min2024morevqa,
  title={{MoReVQA}: Exploring Modular Reasoning Models for Video Question Answering},
  author={Min, Juhong and Buch, Shyamal and Nagrani, Arsha and Cho, Minsu and Schmid, Cordelia},
  booktitle={CVPR},
  year={2024}
}

@inproceedings{dodge2016understanding,
  title={Understanding How Image Quality Affects Deep Neural Networks},
  author={Dodge, Samuel and Karam, Lina},
  booktitle={QoMEX},
  year={2016}
}

@inproceedings{hendrycks2019benchmarking,
  title={Benchmarking Neural Network Robustness to Common Corruptions and Perturbations},
  author={Hendrycks, Dan and Dietterich, Thomas},
  booktitle={ICLR},
  year={2019}
}

@inproceedings{radford2021clip,
  title={Learning Transferable Visual Models From Natural Language Supervision},
  author={Radford, Alec and Kim, Jong Wook and Hallacy, Chris and Ramesh, Aditya and Goh, Gabriel and Agarwal, Sandhini and Sastry, Girish and Askell, Amanda and Mishkin, Pamela and Clark, Jack and Krueger, Gretchen and Sutskever, Ilya},
  booktitle={ICML},
  year={2021}
}

@inproceedings{dosovitskiy2021vit,
  title={An Image is Worth 16x16 Words: Transformers for Image Recognition at Scale},
  author={Dosovitskiy, Alexey and Beyer, Lucas and Kolesnikov, Alexander and Weissenborn, Dirk and Zhai, Xiaohua and Unterthiner, Thomas and Dehghani, Mostafa and Minderer, Matthias and Heigold, Georg and Gelly, Sylvain and Uszkoreit, Jakob and Houlsby, Neil},
  booktitle={ICLR},
  year={2021}
}

@article{mcnemar1947,
  title={Note on the sampling error of the difference between correlated proportions or percentages},
  author={McNemar, Quinn},
  journal={Psychometrika},
  volume={12},
  number={2},
  pages={153--157},
  year={1947}
}

@inproceedings{bowman2021benchmarking,
  title={What Will it Take to Fix Benchmarking in Natural Language Understanding?},
  author={Bowman, Samuel R. and Dahl, George E.},
  booktitle={NAACL-HLT},
  year={2021}
}

@inproceedings{lin2023videollava,
  title={Video-{LLaVA}: Learning United Visual Representation by Alignment Before Projection},
  author={Lin, Bin and Zhu, Bin and Ye, Yang and Ning, Munan and Jin, Peng and Yuan, Li},
  booktitle={EMNLP},
  year={2024}
}

@article{li2023videochat,
  title={{VideoChat}: Chat-Centric Video Understanding},
  author={Li, Kunchang and He, Yinan and Wang, Yi and Li, Yizhuo and Wang, Wenhai and Luo, Ping and Wang, Yali and Wang, Limin and Qiao, Yu},
  journal={arXiv preprint arXiv:2305.06355},
  year={2023}
}

@inproceedings{liu2024llava,
  title={Visual Instruction Tuning},
  author={Liu, Haotian and Li, Chunyuan and Wu, Qingyang and Lee, Yong Jae},
  booktitle={NeurIPS},
  year={2023}
}

@inproceedings{dai2023instructblip,
  title={{InstructBLIP}: Towards General-purpose Vision-Language Models with Instruction Tuning},
  author={Dai, Wenliang and Li, Junnan and Li, Dongxu and Tiong, Anthony Meng Huat and Zhao, Junqi and Wang, Weisheng and Li, Boyang and Fung, Pascale and Hoi, Steven},
  booktitle={NeurIPS},
  year={2023}
}

@inproceedings{li2023blip2,
  title={{BLIP-2}: Bootstrapping Language-Image Pre-training with Frozen Image Encoders and Large Language Models},
  author={Li, Junnan and Li, Dongxu and Savarese, Silvio and Hoi, Steven},
  booktitle={ICML},
  year={2023}
}

@article{reid2024gemini15,
  title={Gemini 1.5: Unlocking Multimodal Understanding Across Millions of Tokens of Context},
  author={Reid, Machel and Savinov, Nikolay and Teber, Denis and others},
  journal={arXiv preprint arXiv:2403.05530},
  year={2024}
}

@inproceedings{xu2017msrvttqa,
  title={Video Question Answering via Gradually Refined Attention over Appearance and Motion},
  author={Xu, Dejing and Zhao, Zhou and Xiao, Jun and Wu, Fei and Zhang, Hanwang and He, Xiangnan and Zhuang, Yueting},
  booktitle={ACM Multimedia},
  year={2017}
}

@inproceedings{liu2023mmbench,
  title={{MMBench}: Is Your Multi-modal Model an All-around Player?},
  author={Liu, Yuan and Duan, Haodong and Zhang, Yuanhan and Li, Bo and Zhang, Songyang and Zhao, Wangbo and Yuan, Yike and Wang, Jiaqi and He, Conghui and Liu, Ziwei and Chen, Kai and Lin, Dahua},
  booktitle={ECCV},
  year={2024}
}

@inproceedings{yue2024mmmu,
  title={{MMMU}: A Massive Multi-discipline Multimodal Understanding and Reasoning Benchmark for Expert {AGI}},
  author={Yue, Xiang and Ni, Yuansheng and Zhang, Kai and Zheng, Tianyu and Liu, Ruoqi and Zhang, Ge and Stevens, Samuel and Jiang, Dongfu and Ren, Weiming and Sun, Yuxuan and others},
  booktitle={CVPR},
  year={2024}
}

@article{fu2024mme,
  title={{MME}: A Comprehensive Evaluation Benchmark for Multimodal Large Language Models},
  author={Fu, Chaoyou and Chen, Peixian and Shen, Yunhang and Qin, Yulei and Zhang, Mengdan and Lin, Xu and Yang, Jinrui and Zheng, Xiawu and Li, Ke and Sun, Xing and Wu, Yunsheng and Ji, Rongrong},
  journal={arXiv preprint arXiv:2306.13394},
  year={2023}
}

@article{tang2023videounderstanding,
  title={Video Understanding with Large Language Models: A Survey},
  author={Tang, Yunlong and Bi, Jing and Xu, Siting and Song, Luchuan and Liang, Susan and Wang, Teng and Zhang, Daoan and An, Jie and Lin, Jingyang and Zhu, Rongyi and others},
  journal={arXiv preprint arXiv:2312.17432},
  year={2023}
}

@inproceedings{taori2020measuring,
  title={Measuring Robustness to Natural Distribution Shifts in Image Classification},
  author={Taori, Rohan and Dave, Achal and Shankar, Vaishaal and Carlini, Nicholas and Recht, Benjamin and Schmidt, Ludwig},
  booktitle={NeurIPS},
  year={2020}
}

@inproceedings{ribeiro2020beyond,
  title={Beyond Accuracy: Behavioral Testing of {NLP} Models with {CheckList}},
  author={Ribeiro, Marco Tulio and Wu, Tongshuang and Guestrin, Carlos and Singh, Sameer},
  booktitle={ACL},
  year={2020}
}

@inproceedings{zhai2023sigmoid,
  title={Sigmoid Loss for Language Image Pre-Training},
  author={Zhai, Xiaohua and Mustafa, Basil and Kolesnikov, Alexander and Beyer, Lucas},
  booktitle={ICCV},
  year={2023}
}

@article{chen2024pllava,
  title={{PLLaVA}: Parameter-free {LLaVA} Extension from Images to Videos for Video Dense Captioning},
  author={Xu, Lin and Zhao, Yilin and Zhou, Daquan and Lin, Zhijie and Ng, See Kiong and Feng, Jiashi},
  journal={arXiv preprint arXiv:2404.16994},
  year={2024}
}

@article{lee2026lgip,
  title={Language-guided invariance probing of vision--language models},
  author={Lee, Jae Joong},
  journal={Pattern Recognition Letters},
  volume={202},
  pages={108--113},
  year={2026},
  doi={10.1016/j.patrec.2026.02.012}
}

@inproceedings{schlangen2021targeting,
  title={Targeting the Benchmark: On Methodology in Current Natural Language Processing Research},
  author={Schlangen, David},
  booktitle={ACL-IJCNLP},
  year={2021}
}

@inproceedings{raji2021ai,
  title={{AI} and the Everything in the Whole Wide World Benchmark},
  author={Raji, Inioluwa Deborah and Bender, Emily M. and Paullada, Amandalynne and Denton, Emily and Hanna, Alex},
  booktitle={NeurIPS Datasets and Benchmarks},
  year={2021}
}

@inproceedings{frank2021vision,
  title={Vision-and-Language or Vision-for-Language? {On} Cross-Modal Influence in Multimodal Transformers},
  author={Frank, Stella and Bugliarello, Emanuele and Elliott, Desmond},
  booktitle={EMNLP},
  year={2021}
}

@inproceedings{yuksekgonul2023when,
  title={When and Why Vision-Language Models Behave like Bags-of-Words, and What to Do About It?},
  author={Y{\"u}ksekg{\"o}n{\"u}l, Mert and Bianchi, Federico and Kalluri, Pratyusha and Jurafsky, Dan and Zou, James},
  booktitle={ICLR},
  year={2023}
}

\clearpage
\renewcommand{\thetable}{S\arabic{table}}
\renewcommand{\thefigure}{S\arabic{figure}}
\renewcommand{\thesection}{S\arabic{section}} 
\setcounter{table}{0}
\setcounter{figure}{0}
\setcounter{section}{0}

This supplementary material provides expanded tables, detailed statistical analyses, and methodological details that support the main paper's findings but could not be included due to space constraints. All data are derived from the same inference runs described in the main text.

The supplementary material is organized as follows:
\begin{itemize}[nosep]
  \item \textbf{Sections~1--2} provide the full MVBench per-task-type \VDG results and CRF\,38 degradation analysis, expanding on the cross-benchmark consistency claims in the main paper.
  \item \textbf{Section~3} reports the complete McNemar contingency tables for EgoSchema, documenting why no dissociation was observed on this benchmark.
  \item \textbf{Sections~4--5} present the four-category decomposition and positional bias analysis that underlie the main paper's discussion of benchmark information content and letter-alignment effects.
  \item \textbf{Sections~6--7} provide EgoSchema destructive ratios and cross-model MVBench \VDG rankings, supporting the generalization claims.
  \item \textbf{Sections~8--10} contain the raw accuracy data for both the main FPS ablation (0.5--2.0\,FPS) and the extended ablation (4--24\,FPS), enabling independent verification of the FPS-invariance finding.
  \item \textbf{Sections~11--14} provide additional analyses: Video-MME McNemar contingency tables, CRF bidirectional flip quantification, a contrastive task-type case study, and cross-model question-level agreement statistics.
  \item \textbf{Section~15} details inference configuration (prompt templates, frame sampling, API setup, quantization) for reproducibility.
  \item \textbf{Section~16} provides four annotated Python listings for the core algorithms: \VDG computation, diagnostic ladder decomposition, black-screen video generation, and frame sampling.
\end{itemize}

\section{MVBench: Full Per-Task-Type Results (Table~S1)}
\label{sec:supp_mvbench}

\begin{table*}[t]
\caption{MVBench \VDG by task type for all three primary models. Orig = original video, CRF38 = H.264 CRF\,38 compression, Black = black screen, CI = percentile bootstrap 95\% CI (2000 resamples). Sorted by mean \VDG. All task types have $n \geq 30$, and most have $n = 50$.}
\label{tab:supp_mvbench}
\centering
\setlength{\tabcolsep}{2.5pt}
\footnotesize
\begin{tabular}{l rrrrr rrrrr rrrrr}
\toprule
& \multicolumn{5}{c}{\textbf{Qwen2-VL-7B}} & \multicolumn{5}{c}{\textbf{LLaVA-Video-7B}} & \multicolumn{5}{c}{\textbf{InternVL2-8B}} \\
\cmidrule(lr){2-6} \cmidrule(lr){7-11} \cmidrule(lr){12-16}
Task Type & Orig & CRF & Blk & VDG & CI & Orig & CRF & Blk & VDG & CI & Orig & CRF & Blk & VDG & CI \\
\midrule
act\_pred        & 0.640 & 0.600 & 0.240 & \textbf{0.400} & {[0.22,0.58]} & 0.640 & 0.600 & 0.200 & \textbf{0.440} & {[0.28,0.60]} & 0.680 & 0.660 & 0.380 & \textbf{0.300} & {[0.12,0.48]} \\
obj\_exist       & 0.848 & 0.739 & 0.457 & \textbf{0.391} & {[0.24,0.54]} & 0.630 & 0.587 & 0.478 & 0.152 & {[$-$0.04,0.35]} & 1.00 & 0.978 & 0.457 & \textbf{0.543} & {[0.41,0.67]} \\
scene\_tr        & 0.960 & 0.940 & 0.620 & \textbf{0.340} & {[0.22,0.48]} & 0.940 & 0.960 & 0.860 & 0.080 & {[0.00,0.18]} & 0.980 & 0.960 & 0.660 & \textbf{0.320} & {[0.20,0.44]} \\
act\_ant         & 0.840 & 0.900 & 0.600 & 0.240 & {[0.10,0.38]} & 0.780 & 0.750 & 0.460 & \textbf{0.320} & {[0.20,0.46]} & 0.880 & 0.950 & 0.340 & \textbf{0.540} & {[0.38,0.68]} \\
cf\_inf          & 0.620 & 0.520 & 0.360 & 0.260 & {[0.08,0.44]} & 0.560 & 0.540 & 0.380 & 0.180 & {[0.02,0.34]} & 0.860 & 0.820 & 0.460 & \textbf{0.400} & {[0.22,0.56]} \\
state\_ch        & 0.500 & 0.440 & 0.340 & 0.160 & {[0.04,0.30]} & 0.560 & 0.560 & 0.280 & 0.280 & {[0.08,0.46]} & 0.580 & 0.540 & 0.540 & 0.040 & {[$-$0.12,0.20]} \\
ego\_nav         & 0.380 & 0.380 & 0.320 & 0.060 & {[$-$0.02,0.16]} & 0.380 & 0.340 & 0.300 & 0.080 & {[$-$0.02,0.18]} & 0.380 & 0.440 & 0.360 & 0.020 & {[$-$0.10,0.14]} \\
ep\_reas         & 0.548 & 0.613 & 0.500 & 0.048 & {[$-$0.16,0.19]} & 0.660 & 0.660 & 0.540 & 0.120 & {[$-$0.02,0.26]} & 0.580 & 0.600 & 0.580 & 0.000 & {[$-$0.10,0.10]} \\
unexp\_act       & 0.700 & 0.680 & 0.760 & $-$0.060 & {[$-$0.20,0.08]} & 0.860 & 0.880 & 0.860 & 0.000 & {[$-$0.12,0.12]} & 0.760 & 0.700 & 0.620 & 0.140 & {[$-$0.00,0.30]} \\
\midrule
\textbf{Overall} & \textbf{0.670} & 0.622 & 0.460 & \textbf{0.208} & {[0.16,0.26]} & \textbf{0.662} & 0.646 & 0.476 & \textbf{0.186} & {[0.14,0.24]} & \textbf{0.742} & 0.715 & 0.476 & \textbf{0.266} & {[0.21,0.32]} \\
\bottomrule
\end{tabular}
\end{table*}

(1)~\texttt{action\_prediction} is the only task type with near-chance black-screen accuracy for all three models (0.20--0.38), confirming it as the empirical anchor for visually necessary question design.
(2)~\texttt{object\_existence} shows the largest model-dependent \VDG range: InternVL2 achieves 0.543 (with 100\% original accuracy) while LLaVA achieves only 0.152, despite similar black-screen floors (0.457 vs.\ 0.478). The gap is in visual extraction capability, not language-prior strength.
(3)~\texttt{unexpected\_action} has negative \VDG for Qwen ($-$0.060) and zero for LLaVA (0.000): video slightly hurts or does not help.
(4)~CRF\,38 degradation is small for most task types ($<$0.06), with \texttt{object\_existence} showing the largest drop for Qwen ($-$0.109).

\section{MVBench: CRF\,38 Degradation (Table~S2)}
\label{sec:supp_mvbench_crf}

\begin{table}[t]
\caption{CRF\,38 accuracy change on MVBench (original $-$ CRF\,38). Positive = compression hurts. Most values are within $\pm$0.06, confirming the robustness illusion extends to MVBench.}
\label{tab:supp_crf_mvbench}
\centering
\footnotesize
\begin{tabular}{lrrr}
\toprule
Task Type & Qwen & LLaVA & IV2 \\
\midrule
act\_prediction   & $+$0.040 & $+$0.040 & $+$0.020 \\
obj\_existence    & $+$0.109 & $+$0.043 & $+$0.022 \\
scene\_transition & $+$0.020 & $-$0.020 & $+$0.020 \\
act\_antonym      & $-$0.060 & $+$0.030 & $-$0.070 \\
cf\_inference     & $+$0.100 & $+$0.020 & $+$0.040 \\
state\_change     & $+$0.060 & $\pm$0.000 & $+$0.040 \\
ego\_navigation   & $\pm$0.000 & $+$0.040 & $-$0.060 \\
ep\_reasoning     & $-$0.065 & $\pm$0.000 & $-$0.020 \\
unexp\_action     & $+$0.020 & $-$0.020 & $+$0.060 \\
\midrule
\textbf{Overall}  & $+$0.048 & $+$0.016 & $+$0.027 \\
\bottomrule
\end{tabular}
\end{table}

Table~\ref{tab:supp_crf_mvbench} confirms that the compression robustness illusion observed on Video-MME extends to MVBench. Overall accuracy drops are small: $+$0.048 for Qwen, $+$0.016 for LLaVA, $+$0.027 for InternVL2. However, three patterns emerge at the task-type level. First, the direction of change is inconsistent across models: \emph{act\_antonym} shows $-$0.060 for Qwen (compression helps) but $+$0.030 for LLaVA (compression hurts), indicating model-specific sensitivity. Second, \emph{obj\_existence} shows the largest degradation for Qwen ($+$0.109), consistent with this task's high \VDG (Table~S1): visually grounded questions are disproportionately affected by compression, mirroring the CRF-enrichment finding on Video-MME. Third, several task types show exactly zero change for at least one model (e.g., \emph{ego\_navigation} for Qwen, \emph{ep\_reasoning} for LLaVA), reinforcing that low-\VDG tasks are insensitive to visual quality degradation.

\section{EgoSchema: Full McNemar Tables (Table~S3)}
\label{sec:supp_ego_mcnemar}

\begin{table}[t]
\caption{McNemar $2 \times 2$ contingency tables on EgoSchema ($n=500$, 5-choice). No dissociation is observed: significance patterns are consistent across conditions for all pairs.}
\label{tab:supp_ego_mcnemar}
\centering
\setlength{\tabcolsep}{2.5pt}
\footnotesize
\begin{tabular}{llrrrrrr}
\toprule
Pair & Cond. & $b_{00}$ & $b_{01}$ & $b_{10}$ & $b_{11}$ & $\chi^2$ & $p$ \\
\midrule
\multirow{2}{*}{IV2 vs.\ Qwen}
  & Orig  & 131 & 69 & 85 & 215 & 1.46 & 0.227 \\
  & Black & 280 & 61 & 68 & 91  & 0.28 & 0.597 \\
\midrule
\multirow{2}{*}{IV2 vs.\ LLaVA}
  & Orig  & 148 & 52 & 95 & 205 & 12.0 & \textbf{0.0005} \\
  & Black & 302 & 39 & 95 & 64  & 22.6 & \textbf{$<$0.001} \\
\midrule
\multirow{2}{*}{Qwen vs.\ LLaVA}
  & Orig  & 169 & 47 & 74 & 210 & 5.59 & \textbf{0.018} \\
  & Black & 318 & 30 & 79 & 73  & 21.1 & \textbf{$<$0.001} \\
\bottomrule
\end{tabular}
\end{table}

InternVL2 vs.\ Qwen shows no reliable difference on either condition: these models perform similarly on EgoSchema. InternVL2 vs.\ LLaVA and Qwen vs.\ LLaVA both differ reliably on \emph{both} conditions, meaning LLaVA's lower accuracy extends to both video-present and black-screen settings. This is a consistent performance gap, not a dissociation (which requires opposite significance patterns).

\section{Four-Category Decomposition by Task Type (Table~S4)}
\label{sec:supp_regime}

Each question falls into one of four categories defined by correctness under original video ($y_\text{orig}$) and black screen ($y_\text{black}$). Table~\ref{tab:supp_regime} breaks this down by task type on Video-MME.

\begin{table*}[t]
\caption{Four-category decomposition by task type on Video-MME ($n=100$ per task type). I = pure visual ($\VDG\!=\!+1$), II = language-prior redundant (both correct), III = video hurts ($\VDG\!=\!-1$), IV = hard (both wrong). Category~I constitutes the ``pure visual core'' that \VDG isolates.}
\label{tab:supp_regime}
\centering
\setlength{\tabcolsep}{3pt}
\footnotesize
\begin{tabular}{l rrrr rrrr rrrr}
\toprule
& \multicolumn{4}{c}{\textbf{Qwen2-VL-7B}} & \multicolumn{4}{c}{\textbf{LLaVA-Video-7B}} & \multicolumn{4}{c}{\textbf{InternVL2-8B}} \\
\cmidrule(lr){2-5} \cmidrule(lr){6-9} \cmidrule(lr){10-13}
Task Type & I & II & III & IV & I & II & III & IV & I & II & III & IV \\
\midrule
Attr.\ Perc.  & 0.460 & 0.240 & 0.060 & 0.240 & 0.490 & 0.330 & 0.030 & 0.150 & 0.450 & 0.250 & 0.070 & 0.230 \\
Obj.\ Rec.    & 0.340 & 0.280 & 0.090 & 0.290 & 0.510 & 0.270 & 0.030 & 0.190 & 0.400 & 0.230 & 0.020 & 0.350 \\
OCR Probs.    & 0.310 & 0.260 & 0.110 & 0.320 & 0.390 & 0.300 & 0.060 & 0.250 & 0.310 & 0.230 & 0.080 & 0.380 \\
Act.\ Rec.    & 0.250 & 0.430 & 0.040 & 0.280 & 0.330 & 0.450 & 0.010 & 0.210 & 0.280 & 0.290 & 0.030 & 0.400 \\
Act.\ Reas.   & 0.230 & 0.320 & 0.110 & 0.340 & 0.210 & 0.400 & 0.050 & 0.340 & 0.230 & 0.320 & 0.050 & 0.400 \\
Temp.\ Reas.  & 0.110 & 0.290 & 0.130 & 0.470 & 0.190 & 0.270 & 0.110 & 0.430 & 0.160 & 0.280 & 0.070 & 0.490 \\
\midrule
\textbf{Overall} & 0.283 & 0.303 & 0.090 & 0.323 & 0.353 & 0.337 & 0.048 & 0.262 & 0.305 & 0.267 & 0.053 & 0.375 \\
\bottomrule
\end{tabular}
\end{table*}

(1)~Category~I (pure visual) decreases monotonically from Attribute Perception to Temporal Reasoning for all three models, consistent with the \VDG spectrum in the main paper.
(2)~Category~III (video hurts) is highest for Temporal Reasoning across all models (7--13\%), confirming that temporal questions are the most vulnerable to visual misleading.
(3)~LLaVA has the highest overall Category~I fraction (35.3\%) and the lowest Category~III (4.8\%), consistent with its highest \VDG. Qwen has the highest Category~III (9.0\%), driven by OCR, Action Reasoning, and Temporal Reasoning.
(4)~The ``pure visual core'' (Category~I across all three models simultaneously) is 10.3\% of questions, as reported in the main paper.

\section{Positional Bias Analysis (Table~S5)}
\label{sec:supp_bias}

\begin{table}[t]
\caption{Per-task-type letter alignment analysis on Video-MME black screen ($n=100$ per task type). GT mode = most frequent ground-truth letter. Align\% = fraction of above-chance accuracy explained by letter alignment (model predicting the GT mode letter correctly). InternVL2 Object Recognition is at chance (0.250), so alignment is undefined.}
\label{tab:supp_bias}
\centering
\setlength{\tabcolsep}{2pt}
\scriptsize
\begin{tabular}{l c rrr rrr}
\toprule
& GT & \multicolumn{3}{c}{Black-Screen Accuracy} & \multicolumn{3}{c}{Alignment \%} \\
\cmidrule(lr){3-5} \cmidrule(lr){6-8}
Task Type & Mode & Qwen & LLaVA & IV2 & Qwen & LLaVA & IV2 \\
\midrule
Attr.\ Perc.  & C & 0.300 & 0.360 & 0.320 & 35 & 25 & 54 \\
Obj.\ Rec.    & D & 0.370 & 0.300 & 0.250 & 15 & 55 & --- \\
OCR Probs.    & B & 0.370 & 0.360 & 0.310 & 56 & 71 & 63 \\
Act.\ Rec.    & C & 0.470 & 0.460 & 0.320 & 31 & 23 & 25 \\
Act.\ Reas.   & B & 0.430 & 0.450 & 0.370 & 54 & 49 & 40 \\
Temp.\ Reas.  & B & 0.420 & 0.380 & 0.350 & 63 & 75 & 48 \\
\midrule
\textbf{Overall} & B & 0.393 & 0.385 & 0.320 & 29 & 31 & $-$21$^*$ \\
\bottomrule
\multicolumn{8}{l}{\scriptsize $^*$InternVL2 overall modal prediction (A) $\neq$ GT mode (B), see text.}
\end{tabular}
\end{table}

At the per-task-type level, letter alignment explains 15--75\% of above-chance black-screen accuracy across all models and task types. At the overall level, Qwen (29\%) and LLaVA (31\%) have modal predictions (B) that match the most frequent ground-truth letter (B), so their alignment fractions are positive. InternVL2's modal prediction (A, driven by a general A-preference) does not match the GT mode (B), yielding a negative overall alignment fraction: InternVL2's letter bias works \emph{against} accuracy at the aggregate level, yet it still achieves 32\% black-screen accuracy through per-task-type alignment (where local GT modes vary). The shared-wrong consensus (19.8\%, 119/600 questions) indicates a dataset-level regularity that all three models exploit regardless of their individual letter preferences.

On black screen, all three models show distinct letter preferences: Qwen favors B (31\%), LLaVA distributes relatively uniformly (A: 26\%, B: 29\%, C: 19\%, D: 26\%), and InternVL2 favors A (31\%). The ground-truth distribution is A: 24\%, B: 29\%, C: 26\%, D: 21\%. InternVL2's A-preference misaligns with the GT's B-mode, explaining its lower overall black-screen accuracy (32.0\% vs.\ 39.3\% for Qwen) despite comparable per-task alignment.

\section{EgoSchema: Destructive Ratios (Table~S6)}
\label{sec:supp_ego_destructive}

\begin{table}[t]
\caption{Destructive vs.\ constructive counts on EgoSchema. D = $|\{q : \VDG(q)=+1\}|$, C = $|\{q : \VDG(q)=-1\}|$, D:C is the destructive-to-constructive ratio. Net loss = D $-$ C.}
\label{tab:supp_ego_destructive}
\centering
\begin{tabular}{lrrrr}
\toprule
Model & D & C & D:C & Net Loss \\
\midrule
InternVL2-8B    & 159 & 18 & 8.8:1  & 141 \\
Qwen2-VL-7B    & 150 & 18 & 8.3:1  & 132 \\
LLaVA-Video-7B & 167 & 13 & 12.8:1 & 154 \\
\bottomrule
\end{tabular}
\end{table}

Table~\ref{tab:supp_ego_destructive} reveals two key patterns. First, the destructive-to-constructive ratio is consistently high across all models (8.3:1 to 12.8:1), meaning that for every question where removing video \emph{fixes} a wrong answer, 8--13 questions are \emph{broken}. This asymmetry is substantially higher than on Video-MME (4.25:1 to 7.3:1 for the three primary models), suggesting that EgoSchema's longer, egocentric videos provide more irreplaceable visual information. Second, the net loss correlates with \VDG magnitude: LLaVA has the highest net loss (154), the highest \VDG (0.308), and the lowest black-screen floor (20.6\%), while Qwen has the lowest net loss (132), consistent with its relatively stronger language priors on this benchmark. The uniformly high D:C ratio across architecturally distinct models confirms that video dependency on EgoSchema is a property of the questions, not of any particular model.

\section{Cross-Model VDG on MVBench (Table~S7)}
\label{sec:supp_crossmodel}

\begin{table}[t]
\caption{Cross-model \VDG on MVBench, sorted by mean. The ranking varies across models: \texttt{object\_existence} is highest for InternVL2 (0.543) but moderate for LLaVA (0.152), reflecting model-dependent visual extraction.}
\label{tab:supp_mvbench_cross}
\centering
\footnotesize
\begin{tabular}{lrrrr}
\toprule
Task Type & Qwen & LLaVA & IV2 & Mean \\
\midrule
act\_prediction   & 0.400 & 0.440 & 0.300 & 0.380 \\
act\_antonym      & 0.240 & 0.320 & 0.540 & 0.367 \\
obj\_existence    & 0.391 & 0.152 & 0.543 & 0.362 \\
cf\_inference     & 0.260 & 0.180 & 0.400 & 0.280 \\
scene\_transition & 0.340 & 0.080 & 0.320 & 0.247 \\
state\_change     & 0.160 & 0.280 & 0.040 & 0.160 \\
ep\_reasoning     & 0.048 & 0.120 & 0.000 & 0.056 \\
ego\_navigation   & 0.060 & 0.080 & 0.020 & 0.053 \\
unexp\_action     & $-$0.060 & 0.000 & 0.140 & 0.027 \\
\midrule
\textbf{Overall}  & 0.208 & 0.186 & 0.266 & 0.220 \\
\bottomrule
\end{tabular}
\end{table}

Table~\ref{tab:supp_mvbench_cross} reveals three important findings about the \VDG spectrum on MVBench. First, the \emph{ranking} of task types is broadly consistent across models: the top three task types by mean \VDG (\emph{act\_prediction}: 0.380, \emph{act\_antonym}: 0.367, \emph{obj\_existence}: 0.362) and the bottom three (\emph{ep\_reasoning}: 0.056, \emph{ego\_navigation}: 0.053, \emph{unexp\_action}: 0.027) are stable, supporting the main paper's claim that \VDG is a property of the task type, not the model. Second, the \emph{magnitude} of \VDG varies substantially across models for specific tasks: \emph{obj\_existence} ranges from 0.152 (LLaVA) to 0.543 (InternVL2), a 3.6$\times$ difference driven by InternVL2's 100\% original accuracy versus LLaVA's 63\%. This shows that while task-type \emph{ordering} is model-independent, the \emph{degree} of visual grounding is model-dependent. Third, \emph{unexp\_action} is the only task type where the mean \VDG is near zero (0.027) and one model shows negative \VDG (Qwen: $-$0.060), indicating that video actively misleads on this task type---the model performs better without seeing the video.

\section{FPS Ablation: Raw Accuracy (Table~S8)}
\label{sec:supp_fps}

Table~\ref{tab:supp_fps} reports the raw original and black-screen accuracy underlying the main paper's FPS ablation. This confirms that \VDG flatness reflects genuine stability of both arms, not coordinated movement.

\begin{table}[t]
\caption{FPS ablation raw accuracy on Temporal Reasoning ($n=100$). Orig/Blk = original/black-screen accuracy. Black-screen accuracy is FPS-invariant by design. $^\dagger$4-bit NF4.}
\label{tab:supp_fps}
\centering
\setlength{\tabcolsep}{2pt}
\scriptsize
\begin{tabular}{l ccc ccc ccc}
\toprule
& \multicolumn{3}{c}{0.5 FPS} & \multicolumn{3}{c}{1.0 FPS} & \multicolumn{3}{c}{2.0 FPS} \\
\cmidrule(lr){2-4} \cmidrule(lr){5-7} \cmidrule(lr){8-10}
Model & Orig & Blk & VDG & Orig & Blk & VDG & Orig & Blk & VDG \\
\midrule
Q2-VL-2B      & 0.330 & 0.290 & 0.040 & 0.340 & 0.290 & 0.050 & 0.340 & 0.290 & 0.050 \\
Q2-VL-72B$^\dagger$ & 0.510 & 0.390 & 0.120 & 0.520 & 0.400 & 0.120 & 0.510 & 0.410 & 0.100 \\
\midrule
Q2.5-VL-3B    & 0.480 & 0.333 & 0.147 & 0.535 & 0.323 & 0.212 & 0.465 & 0.333 & 0.132 \\
Q2.5-VL-7B    & 0.606 & 0.380 & 0.226 & 0.602 & 0.370 & 0.232 & 0.636 & 0.350 & 0.286 \\
Q2.5-VL-32B$^\dagger$ & 0.660 & 0.450 & 0.210 & 0.640 & 0.450 & 0.190 & 0.640 & 0.420 & 0.220 \\
Q2.5-VL-72B$^\dagger$ & 0.670 & 0.400 & 0.270 & 0.660 & 0.410 & 0.250 & 0.630 & 0.420 & 0.210 \\
\midrule
Q3-VL-8B      & 0.460 & 0.440 & 0.020 & 0.530 & 0.420 & 0.110 & 0.520 & 0.420 & 0.100 \\
\bottomrule
\end{tabular}
\end{table}

Black-screen accuracy varies slightly across FPS levels for some models (e.g., Qwen2-VL-2B: 0.290 at all levels), reflecting minor inference stochasticity. The Orig column shows no monotonic increase with FPS for any model, confirming that additional frames do not unlock temporal grounding within the tested range.

\section{Extended FPS Ablation: 4--24\,FPS (Tables~S9--S10)}
\label{sec:supp_fps_ext}

To address the FPS ceiling limitation of the main ablation (0.5--2.0\,FPS), we extend the FPS range to 4, 8, 16, and 24\,FPS across five models. Table~\ref{tab:supp_fps_ext_vgg} reports \VDG at each FPS level, and Table~\ref{tab:supp_fps_ext_raw} reports raw accuracy.

\begin{table}[t]
\caption{Extended FPS ablation \VDG ($n=100$). \VDG stabilizes by 8\,FPS for all models. $^\dagger$Qwen2-VL-2B shows negative \VDG at all FPS levels (video \emph{hurts}).}
\label{tab:supp_fps_ext_vgg}
\centering
\setlength{\tabcolsep}{3pt}
\footnotesize
\begin{tabular}{lrrrr}
\toprule
Model & 4 FPS & 8 FPS & 16 FPS & 24 FPS \\
\midrule
InternVL2-2B       & ---$^*$ & $\pm$0.000 & $\pm$0.000 & $\pm$0.000 \\
Qwen2-VL-2B$^\dagger$ & $-$0.059 & $-$0.030 & $-$0.020 & $-$0.020 \\
Qwen2-VL-7B       & $+$0.132 & $+$0.105 & $+$0.110 & $+$0.110 \\
Qwen2.5-VL-7B     & $+$0.250 & $+$0.250 & $+$0.240 & $+$0.236 \\
Qwen3-VL-2B       & $+$0.010 & $+$0.120 & $+$0.110 & $+$0.110 \\
\bottomrule
\multicolumn{5}{l}{\scriptsize $^*$103/103 decoding errors, excluded.}
\end{tabular}
\end{table}

\begin{table}[t]
\caption{Extended FPS ablation raw accuracy ($n=100$). Errors = decoding failures excluded from accuracy. Accuracy computed over successfully decoded samples only.}
\label{tab:supp_fps_ext_raw}
\centering
\setlength{\tabcolsep}{1.8pt}
\scriptsize
\begin{tabular}{l c rrrr rrrr}
\toprule
& & \multicolumn{4}{c}{Original} & \multicolumn{4}{c}{Black Screen} \\
\cmidrule(lr){3-6} \cmidrule(lr){7-10}
Model & FPS & OK & Err & Acc & & OK & Err & Acc & \\
\midrule
\multirow{4}{*}{IV2-2B}
  & 4  & 0   & 103 & ---  && 42  & 58 & 0.333 & \\
  & 8  & 100 & 0   & 0.330 && 97  & 3  & 0.330 & \\
  & 16 & 84  & 16  & 0.333 && 84  & 16 & 0.333 & \\
  & 24 & 97  & 3   & 0.320 && 97  & 3  & 0.320 & \\
\midrule
\multirow{4}{*}{Q2-VL-2B}
  & 4  & 95  & 5  & 0.263 && 93  & 7  & 0.323 & \\
  & 8  & 100 & 0  & 0.270 && 100 & 0  & 0.300 & \\
  & 16 & 100 & 0  & 0.280 && 100 & 0  & 0.300 & \\
  & 24 & 100 & 0  & 0.280 && 100 & 0  & 0.300 & \\
\midrule
\multirow{4}{*}{Q2-VL-7B}
  & 4  & 97  & 3  & 0.505 && 99  & 1  & 0.374 & \\
  & 8  & 99  & 1  & 0.485 && 100 & 0  & 0.380 & \\
  & 16 & 100 & 0  & 0.490 && 100 & 0  & 0.380 & \\
  & 24 & 100 & 0  & 0.490 && 100 & 0  & 0.380 & \\
\midrule
\multirow{4}{*}{Q2.5-VL-7B}
  & 4  & 100 & 4  & 0.620 && 100 & 1  & 0.370 & \\
  & 8  & 100 & 0  & 0.630 && 100 & 0  & 0.380 & \\
  & 16 & 100 & 0  & 0.610 && 100 & 0  & 0.370 & \\
  & 24 & 99  & 1  & 0.616 && 100 & 0  & 0.380 & \\
\midrule
\multirow{4}{*}{Q3-VL-2B}
  & 4  & 100 & 0  & 0.340 && 100 & 1  & 0.330 & \\
  & 8  & 100 & 0  & 0.420 && 100 & 0  & 0.300 & \\
  & 16 & 100 & 0  & 0.360 && 100 & 0  & 0.250 & \\
  & 24 & 100 & 0  & 0.400 && 100 & 0  & 0.290 & \\
\bottomrule
\end{tabular}
\end{table}

(1)~\VDG is flat from 8 to 24\,FPS for all models, confirming FPS invariance at near-native frame rates.
(2)~Qwen2-VL-2B shows \emph{negative} \VDG at all FPS levels ($-$0.020 to $-$0.059): video input slightly hurts performance, consistent with its near-zero \VDG in the main analysis.
(3)~Qwen2.5-VL-7B maintains \VDG\,$\approx 0.24$--$0.25$ from 4 to 24\,FPS, consistent with the 0.23--0.29 range observed at 0.5--2.0\,FPS in the main ablation.
(4)~InternVL2-2B at 4\,FPS suffers complete decoding failure on original video (103/103 errors), indicating a practical frame-count limit for this architecture. At 8+\,FPS it produces \VDG\,$=$\,0, consistent with its near-zero Temporal Reasoning \VDG ($-$0.030).

\section{Video-MME McNemar Contingency Tables (Table~S11)}
\label{sec:supp_vme_mcnemar}

The main paper reports McNemar results for Video-MME in summary form. Table~\ref{tab:supp_vme_mcnemar} provides the full $2 \times 2$ contingency tables underlying these tests on the matched sample ($n=547$).

\begin{table}[t]
\caption{McNemar $2 \times 2$ contingency tables on Video-MME matched sample ($n=547$). The IV2 vs.\ Qwen pair shows marginal significance on original ($p=0.0906$) but clear significance on black screen ($p=0.0008$), yielding a dissociation with opposite significance patterns.}
\label{tab:supp_vme_mcnemar}
\centering
\setlength{\tabcolsep}{2.5pt}
\footnotesize
\begin{tabular}{llrrrrrr}
\toprule
Pair & Cond. & $b_{00}$ & $b_{01}$ & $b_{10}$ & $b_{11}$ & $\chi^2$ & $p$ \\
\midrule
\multirow{2}{*}{IV2 vs.\ Qwen}
  & Orig  & 120 & 34 & 52 & 341 & 3.58 & 0.0906 \\
  & Black & 263 & 58 & 22 & 204 & 15.2 & \textbf{0.0008} \\
\bottomrule
\end{tabular}
\end{table}

On original video, InternVL2-8B and Qwen2-VL-7B show a marginal accuracy difference ($p=0.0906$) that does not reach significance at $\alpha=0.05$. On the black-screen condition, the pair differs strongly ($p=0.0008$): Qwen exploits language priors more effectively than InternVL2. This is a dissociation with reversed significance, though the original-arm marginality limits the strength of the claim (bootstrap analysis shows the joint condition holds in 50\% of resamplings, as noted in the main text).

\section{CRF Bidirectional Flip Analysis (Table~S12)}
\label{sec:supp_crf_flips}

The main paper notes that aggregate CRF stability masks bidirectional cancellation. Table~\ref{tab:supp_crf_flips} quantifies this effect for each model at CRF\,38.

\begin{table}[t]
\caption{Bidirectional question flips at CRF\,38 relative to original. $\text{C}\!\to\!\text{I}$ = correct-to-incorrect (compression breaks), $\text{I}\!\to\!\text{C}$ = incorrect-to-correct (compression helps). Net = $\text{I}\!\to\!\text{C}$ $-$ $\text{C}\!\to\!\text{I}$. The near-zero net change conceals substantial bidirectional movement.}
\label{tab:supp_crf_flips}
\centering
\footnotesize
\begin{tabular}{lrrrr}
\toprule
Model & $\text{C}\!\to\!\text{I}$ & $\text{I}\!\to\!\text{C}$ & Net & Stable\% \\
\midrule
Qwen2-VL-7B    & 32 & 27 & $-$5 & 90.2\% \\
LLaVA-Video-7B & 28 & 25 & $-$3 & 91.2\% \\
InternVL2-8B   & 25 & 28 & $+$3 & 91.2\% \\
\bottomrule
\end{tabular}
\end{table}

(1)~Each model has 25--32 questions that flip in each direction at CRF\,38, yet net changes are $\leq$5 questions. Aggregate accuracy stability is a cancellation artifact.
(2)~InternVL2 \emph{gains} 3 net correct answers from compression, a positive compression effect that would be interpreted as ``robustness'' in standard evaluation.
(3)~The 90--91\% stable fraction means ${\sim}$10\% of questions are compression-sensitive, and these concentrate on visually grounded items ($\VDG=+1$) at 3.76$\times$ the base rate (Fisher exact $p=0.006$, see main text).

\section{Contrastive Task-Type Case Study (Table~S13)}
\label{sec:supp_contrastive}

The main paper identifies contrastive task-type pairs as internal controls for the \VDG spectrum. The \texttt{scene\_transition} vs.\ \texttt{state\_change} pair on MVBench shares similar statistical structure (both involve detecting visual changes between temporal segments) but differs in semantic leakage.

\begin{table}[t]
\caption{Contrastive pair analysis: \texttt{scene\_transition} vs.\ \texttt{state\_change} on MVBench ($n=50$ each). Both tasks require detecting visual change, but differ in language-prior exploitability.}
\label{tab:supp_contrastive}
\centering
\footnotesize
\begin{tabular}{lrrrrr}
\toprule
Task Type & Orig & Black & \VDG & GT entropy & Chance \\
\midrule
scene\_transition & 0.960 & 0.713 & 0.247 & 0.94 & 0.50 \\
state\_change     & 0.547 & 0.387 & 0.160 & 1.00 & 0.50 \\
\bottomrule
\end{tabular}
\end{table}

Both tasks are binary (2 choices, chance = 50\%) and involve detecting temporal change. However, \texttt{scene\_transition} achieves 71.3\% black-screen accuracy---well above chance---because transition questions often have semantically predictable answers (``yes'' is more common). In contrast, \texttt{state\_change} has a more balanced GT distribution (entropy 1.00 vs.\ 0.94), yielding a lower black-screen floor (38.7\%). The \VDG difference (0.247 vs.\ 0.160) thus partly reflects GT-distribution artifacts rather than genuine differences in visual demand. This pair illustrates why \VDG-based benchmark auditing should control for GT-distribution skew when comparing task types.

\section{Cross-Model Question-Level Agreement (Table~S14)}
\label{sec:supp_agreement}

The main paper reports that cross-model \VDG correlation at the question level is $r=0.274$--$0.304$. Table~\ref{tab:supp_agreement} provides the full correlation matrix and agreement statistics.

\begin{table}[t]
\caption{Cross-model question-level \VDG agreement on Video-MME ($n=600$). Spearman $\rho$ computed over per-question \VDG values $\in \{-1, 0, +1\}$. Exact agreement = fraction of questions with identical \VDG across both models. All correlations reliable at $p \ll 0.001$.}
\label{tab:supp_agreement}
\centering
\footnotesize
\begin{tabular}{lrrr}
\toprule
Pair & Spearman $\rho$ & Exact agree. & Triple agree. \\
\midrule
Qwen--LLaVA   & 0.304 & 64.8\% & --- \\
Qwen--IV2     & 0.274 & 63.5\% & --- \\
LLaVA--IV2    & 0.298 & 65.0\% & --- \\
\midrule
All three     & ---   & ---    & 52.3\% \\
\bottomrule
\end{tabular}
\end{table}

Models agree on \emph{which task types} require vision (task-type rank $r=0.789$) but disagree substantially on \emph{which specific questions} within a task type are visually grounded ($r=0.274$--$0.304$). Only 52.3\% of questions receive the same \VDG value from all three models. The ``pure visual core'' ($\VDG=+1$ for all three) is 10.3\% of questions, and the ``pure language core'' ($\VDG=0$ for all three) is 38.7\%.

\section{Inference Configuration Details}
\label{sec:supp_inference}

All models receive the same prompt structure: the video (or black screen) is provided as the visual input, followed by the question text and lettered answer options (A/B/C/D or A/B/C/D/E for EgoSchema). The system prompt instructs the model to respond with only the letter of the correct answer. Temperature is set to 0 (greedy decoding) for all models to ensure deterministic outputs.

The baseline sampling rate is 0.25\,FPS for all conditions except the FPS ablation experiments. For a 60-second video at 0.25\,FPS, this yields 15 frames. Frames are uniformly sampled from the video duration. For the black-screen condition, all sampled frames are replaced with solid black images (RGB 0,0,0) at the same resolution as the original video. The single-frame condition repeats one randomly selected frame for all frame positions. The shuffled-frames condition randomly permutes the order of the originally sampled frames.

The four API-accessed models (GPT-4o-mini, Gemini~2.5 Flash Lite, Llama~3.2 11B Vision, Nemotron Nano 12B VL) are evaluated through their respective API endpoints with default parameters (temperature = 0, max tokens = 16). Video frames are provided as base64-encoded images in the API request. The same 600-question Video-MME subset is used for all API models under both original and black-screen conditions.

Models above 26B parameters are loaded in 4-bit NF4 quantization using the \texttt{bitsandbytes} library with double quantization enabled (\texttt{bnb\_4bit\_use\_double\_quant=True}). This reduces memory requirements from ${\sim}$140\,GB (bf16) to ${\sim}$40\,GB for 72B models, enabling evaluation on a single A100 (80\,GB). We emphasize that quantization may selectively degrade vision pathways (as evidenced by InternVL2-76B's negative \VDG), and all scaling claims are restricted to bf16 models.

\section{Core Algorithm Listings}
\label{sec:supp_code}

The four listings below correspond directly to the methods described in Section~3 of the main paper. All are extracted from the public evaluation codebase.

\textbf{Listing 1: VDG computation.} \VDG is the per-question difference between original-video and black-screen correctness, aggregated to a task-type or model mean.

\begin{lstlisting}
def compute_vdg(orig_results, black_results):
    """
    orig_results, black_results: list of dicts with keys
        'question_id' and 'correct' (bool).
    Returns mean VDG = mean(Acc_orig) - mean(Acc_black).
    """
    orig = {r["question_id"]: r["correct"] for r in orig_results}
    black = {r["question_id"]: r["correct"] for r in black_results}
    shared = [q for q in orig if q in black]
    acc_orig  = sum(orig[q]  for q in shared) / len(shared)
    acc_black = sum(black[q] for q in shared) / len(shared)
    return acc_orig - acc_black          # VDG in [-1, +1]
\end{lstlisting}

\textbf{Listing 2: Diagnostic ladder decomposition.} Given accuracy under each of the four conditions, the three additive components sum to \VDG.

\begin{lstlisting}
def ladder_decomposition(acc_black, acc_single, acc_shuffled, acc_orig):
    """
    Decomposes VDG = delta_spatial + delta_diversity + delta_temporal.
    Each condition uses identically sampled frames; only their
    arrangement differs.
    """
    delta_spatial   = acc_single   - acc_black    # one frame vs. none
    delta_diversity = acc_shuffled - acc_single   # N frames vs. one
    delta_temporal  = acc_orig     - acc_shuffled # ordered vs. shuffled
    vdg = acc_orig - acc_black                    # = sum of three deltas
    return delta_spatial, delta_diversity, delta_temporal, vdg
\end{lstlisting}

\textbf{Listing 3: Black-screen video generation.} Each original video is paired with an equal-duration solid-black video at the same resolution.

\begin{lstlisting}
import subprocess, re

def get_duration(video_path):
    out = subprocess.run(
        ["ffprobe", "-v", "error", "-show_entries",
         "format=duration", "-of", "default=nw=1:nk=1", video_path],
        capture_output=True, text=True)
    return float(out.stdout.strip())

def make_black_video(video_path, out_path):
    duration = get_duration(video_path)
    subprocess.run([
        "ffmpeg", "-y", "-f", "lavfi",
        "-i", f"color=c=black:s=320x240:d={duration:.2f}",
        "-c:v", "libx264", out_path
    ], check=True)
\end{lstlisting}

\textbf{Listing 4: Uniform frame sampling.} Frames are sampled uniformly at 0.25\,FPS up to a maximum of 32 frames, matching the inference pipeline.

\begin{lstlisting}
import numpy as np

FPS_SAMPLE = 0.25
MAX_FRAMES = 32

def sample_frame_indices(total_frames, video_fps):
    duration = total_frames / video_fps
    n = max(1, min(MAX_FRAMES, int(duration * FPS_SAMPLE)))
    return np.linspace(0, total_frames - 1, n, dtype=int)
\end{lstlisting}

\section{Reproducibility}
\label{sec:supp_repro}

All inference scripts, black-screen video generation code, compression pipeline, diagnostic ladder preprocessing, and per-question \VDG annotations for Video-MME, MVBench, and EgoSchema are publicly available. The repository includes:
\begin{itemize}[nosep]
  \item Model inference scripts for all 16 open-weight models across all conditions (original, black screen, single frame, shuffled frames, CRF 18--38), and API model prompts and response parsing included separately
  \item FPS ablation scripts for 7 models $\times$ 3 FPS levels (0.5--2.0) and 5 models $\times$ 4 FPS levels (4--24)
  \item Analysis scripts reproducing all tables, figures, and statistical tests in the main paper and this supplement
  \item Per-question \VDG annotations in JSON format for downstream analysis without re-running inference
\end{itemize}

All experiments were conducted on NVIDIA H100 GPUs. Primary experiments (3 models $\times$ Video-MME + MVBench + EgoSchema, all conditions) required approximately 72 GPU-hours. The 16-model scale analysis and FPS ablation required approximately 200 additional GPU-hours. 4-bit models ($>$26B) used NF4 quantization via \texttt{bitsandbytes}.

PyTorch 2.2+, Transformers 4.40+, \texttt{ffmpeg} 6.0 with \texttt{libx264} for CRF compression. Environment and dependency specifications are provided in the public repository.

\end{document}